\pdfoutput=1

\documentclass[11pt]{article}
\usepackage{amsmath}
\usepackage{amsmath}
\usepackage{amsfonts}

\usepackage[]{acl}

\usepackage{times}
\usepackage{latexsym}

\usepackage[T1]{fontenc}

\usepackage[utf8]{inputenc}

\usepackage{microtype}
\usepackage{array}
\usepackage{booktabs}
\usepackage{graphicx}
\usepackage{subfigure}
\usepackage{longtable}
\usepackage{multicol}
\usepackage{geometry}
\usepackage{tabularx}
\usepackage{xtab,afterpage}
\usepackage{caption}
\usepackage{times}
\usepackage{latexsym}
\usepackage{amsmath,amssymb,amsfonts}
\usepackage{multirow}
\usepackage{hyperref}
\usepackage{dblfloatfix}
\usepackage{verbatim}
\usepackage{colortbl}
\usepackage{makecell}
\usepackage{xspace}
\usepackage{enumitem}

\newcolumntype{L}[1]{>{\raggedright\let\newline\\\arraybackslash\hspace{0pt}}m{#1}}
\newcolumntype{C}[1]{>{\centering\let\newline\\\arraybackslash\hspace{0pt}}m{#1}}
\newcolumntype{R}[1]{>{\raggedleft\let\newline\\\arraybackslash\hspace{0pt}}m{#1}}

\usepackage{subfiles}
\usepackage{tcolorbox}

\usepackage{inconsolata}

\usepackage{graphicx}

\usepackage{listings}
\usepackage{xcolor}
\usepackage{xcolor}

\title{Disentangling Perception and Reasoning in Multimodal LLMs\\ via Reward Design}

\newcommand{\rFormat}{\ensuremath{\mathcal{R}_{\textsc{format}}}\xspace}
\newcommand{\rOutcome}{\ensuremath{\mathcal{R}_{\textsc{outcome}}}\xspace}
\newcommand{\rStructured}{\ensuremath{\mathcal{R}_{\textsc{structured}}}\xspace}
\newcommand{\rGraded}{\ensuremath{\mathcal{R}_{\textsc{graded}}}\xspace}
\newcommand{\rGrounded}{\ensuremath{\mathcal{R}_{\textsc{grounded}}}\xspace}
\newcommand{\rProcess}{\ensuremath{\mathcal{R}_{\textsc{process}}}\xspace}

\newcommand{\rewardSet}{\{\rFormat, \rOutcome, \rStructured, \rGraded, \rGrounded, \rProcess\}}

\author{Omar Sharif{\textsuperscript{$\spadesuit$}}, Eftekhar Hossain{\textsuperscript{$\clubsuit$}}, Nikhil Singh{\textsuperscript{$\spadesuit$\large{\thanks{\;Equal advising.}}}},  Patrick Ng{\textsuperscript{$\dagger$\large{\footnotemark[1]}}}\\
{\textsuperscript{$\spadesuit$}}Department of Computer Science, Dartmouth College \\
{\textsuperscript{$\clubsuit$}}Department of Computer Science, University of Central Florida\\
\textsuperscript{$\dagger$} Independent Researcher\\
 {\texttt{\{omar.sharif.gr, nikhil.u.singh\}@dartmouth.edu, eftekhar@ucf.edu}}}

\begin{document}
\maketitle

\begin{abstract}


Reinforcement learning with verifiable rewards has driven major gains in LLM reasoning, and it is intuitive to assume this recipe will transfer well to multimodal models. However, multimodal models do two things: first, perceive what is in an image, then reason about what it implies. Because these stages are graded jointly, it is hard to tell how much room reasoning alone has to grow. We study this on algorithmic visual puzzles, where both components are necessary and show that \textit{perception}, not reasoning, is the binding constraint. Replacing images with simple textual descriptions raises performance by over 20 points on average for Claude models. We then evaluate six reward designs aimed at inducing visual grounding during reasoning without chain-of-thought supervision. Training Qwen-2.5-VL-7B with GRPO, reward design induces long, structured reasoning with self-reflection and visual references, yielding a 5.56-point gain over the base model. These gains are, however, uneven; no single reward improves all categories, and rewards with verifiable accuracy signals trade cross-domain transfer for in-domain accuracy. These results point to perception-aware reward design as a path forward, so that signals correct perception at its source rather than the reasoning that inherits its errors.
\end{abstract}






\begin{figure*}[h!]
  \centering
  \includegraphics[width =0.95\linewidth]{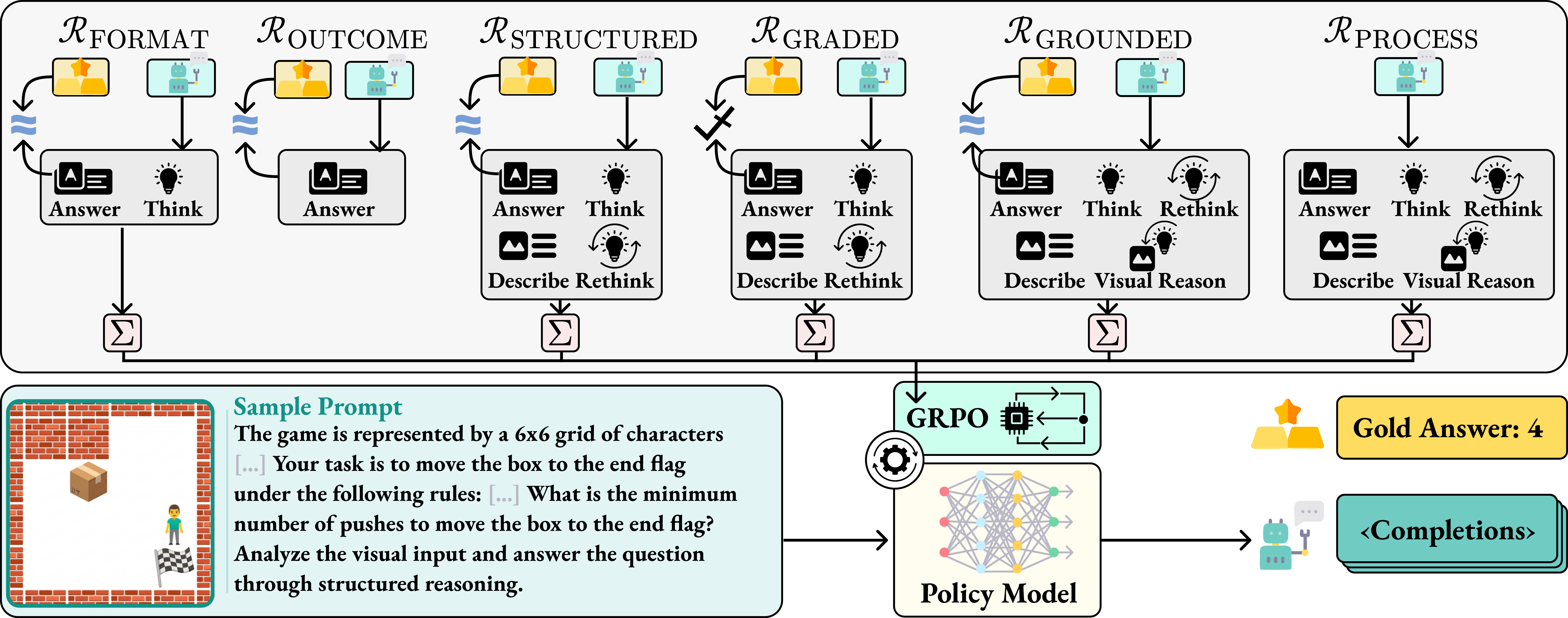}
 \caption{Overview of the GRPO-based training pipeline for MLLMs. For each sample, the policy model generates a completion conditioned on the prompt associated with a specific reward type (\rewardSet). Each completion is then evaluated using the reward calculation framework (\S\ref{sec:reward-functions}). Completions with different rewards are shown in Figure~\ref{completions}. The resulting reward scores are passed into the GRPO trainer, which updates the policy model accordingly.}
 \label{rewards-fig}
\end{figure*}

\begin{table*}[h]
\centering
\small
\renewcommand*{\arraystretch}{0.92}
\rowcolors{2}{gray!10}{cyan!5}
\begin{tabular}{L{2.5cm}L{3cm}L{4cm}L{5.2cm}}
\toprule
\textbf{Reward Function} & \textbf{Reward Content} &  \textbf{Reward Calculation} & \textbf{Design Motivation} \\
\midrule
\rFormat & \texttt{<think, answer>} & $\alpha_t * \text{think} + \alpha_a * \text{accuracy}$ & Baseline reward for RL training for formatted reasoning and accuracy. \\
\midrule
\rOutcome & \texttt{<answer>} & $\alpha_a * \text{accuracy}$ & Is accuracy reward enough to encourage reasoning? Can a model learn the format from the prompt alone, without any separate reward for the format? \\
\midrule
\rStructured & \texttt{<image description, think, rethink, answer>} & $\alpha_i * \text{image} + \alpha_t * \text{think} + \alpha_r * \text{rethink} + \alpha_a * \text{accuracy}$ & Does distribution of reward across intuitive tags incentivize better reasoning? \\
\midrule
\rGraded & \texttt{<image description, think, rethink, answer (partial)>} & $\alpha_i * \text{image} + \alpha_t * \text{think} + \alpha_r * \text{rethink} + \alpha_a * \text{accuracy} (partial)$  & Does giving partial reward to the model when the final answer is close to the ground-truth encourage correct reasoning and answer? \\
\midrule
\rGrounded & \texttt{<think, visual, answer>} & $\alpha_t * \text{think} + \alpha_v * \text{visual} + \alpha_a * \text{accuracy}$ & Focus on integrating visual grounding within reasoning. \\
\midrule
\rProcess & \texttt{<image description, think, rethink, answer-tag>} & $\alpha_i * \text{image} + \alpha_t * \text{think} + \alpha_r * \text{rethink} + \alpha_a * \text{answer-tag}$ & Can the model generalize or improve its performance without any verifiable reward? \\
\hline
\end{tabular}
\caption{Summary of reward functions for incentivizing visual reasoning in MLLMs. \textit{Reward Content} specifies which output tags are evaluated, and \textit{Reward Calculation} indicates the weighting of each reward component. Each non-answer tag’s score is based on the amount of unique content generated within that tag, with counts passed through a $tanh()$ smoothing function to prevent disproportionately large rewards. Parameter values ($\alpha_t, \alpha_r, \alpha_i, \alpha_v, \alpha_a$) are provided in Appendix Table \ref{table:reward_parameters}.}
\label{table:reward_summary}
\end{table*}

\section{Introduction}

We tend to evaluate multimodal models in a simplified way: the answer is right or it's not, and what happened between seeing the problem and writing the answer is effectively a black box. This obscures the fact that there are two steps in multimodal problems: perception and reasoning. For example, given an image of an analog clock and asked what time it is an hour later, the model must first see what is in the image (read the clock), and then reason about it (add an hour). A wrong answer can mean it misread the image, miscomputed from a correct reading, or both. From the answer alone, we cannot immediately tell which one is true. Indeed, while modern such models appear to excel on tasks requiring extensive domain knowledge, they often struggle with tasks that demand visual perception \cite{shojaee2025illusionthinkingunderstandingstrengths}.

A related observation concerns how current MLLMs~\cite{qwen2.5,qwen3}, despite being \textit{trained} to reason in many cases, often produce reasoning that is weakly grounded in the visual input~\cite{fan2025gritteachingmllmsthink}. Indeed, in our own experiments, models frequently address the algorithmic component of a question without explicit reference to the image, or jump directly to an answer with minimal intermediate description. Two failure modes thus appear together: inadequate interpretation of the image, and a tendency to bypass visual grounding altogether in the reasoning chain.

Investigating these failure modes requires both \textbf{(i)} an \textit{evaluation setting} in which perception and reasoning can be disentangled, and \textbf{(ii)} studying \textit{training methods} that can shape both. For (i), we turn to \textbf{visual puzzles}, which require both accurate visual perception and algorithmic reasoning. These have become an effective way to evaluate reasoning capabilities in MLLMs independent of specialized knowledge \cite{chia-etal-2024-puzzlevqa,ghosal-etal-2025-algopuzzlevqa}, showing that such models consistently underperform relative to other reasoning-oriented tasks \cite{song2025visualpuzzlesdecouplingmultimodalreasoning}. They thus provide a useful setting for asking where in the pipeline performance breaks down. In other words, we ask \textbf{what is the bottleneck of MLLMs for solving visual puzzles?} Then, to separate the contributions of perception and reasoning to this gap, we hold the reasoning task constant and vary only the perceptual demand.

For (ii), we turn to reinforcement learning motivated by its empirical success in encouraging effective reasoning in text-only models. For example, generating longer textual reasoning trajectories has been shown to enable self-reflection and improve accuracy~\cite{deepseekai2025deepseekr1incentivizingreasoningcapability}. The central tool for incentivizing this behavior is the \textit{reward function}~\cite{shao2025spuriousrewardsrethinkingtraining}, and this is thus also the primary lever we use to investigate the possibilities of improving the perception--reasoning relationship. This is in contrast to recent attempts to encourage strong multimodal reasoning, which have relied instead on external supervision~\cite{zheng2025deepeyesincentivizingthinkingimages,cao2025groundr1incentivizinggroundedvisual}, typically in the form of annotated bounding boxes, tool outputs, or human-curated reasoning chains. \textbf{We seek a more general, supervision-free solution as have worked for text-based LLMs.} Once again, complex visual puzzles help to ground this goal in a concrete test. In these settings, such annotations are both costly to produce and difficult to define, as a single puzzle may admit multiple valid reasoning paths. The design space of multimodal reward functions is still an emerging area~\cite{su2025thinkingimagesmultimodalreasoning}.

To better understand the impact of rewards on multimodal reasoning models, \textbf{we design and evaluate six candidate reward functions} which systematically vary what information we condition on: \rewardSet~(each is defined in Table~\ref{table:reward_summary}). These variants differ along complementary axes, such as whether the final answer must exactly match the ground truth or whether near-correct answers receive partial credit; whether intermediate visual descriptions are scored separately from the reasoning chain; and whether ground-truth supervision is provided at all. We train Qwen-2.5-VL-7B~\cite{qwen2.5} with each reward under GRPO~\cite{shao2024deepseekmathpushinglimitsmathematical}, and evaluate on both in-domain and cross-domain puzzle categories. Figure~\ref{rewards-fig} illustrates the training pipeline.

Our results show that these post-trained models exhibit qualitatively different reasoning behavior from the base model. Under reward variants that score intermediate image descriptions, the model generates extended, structured reasoning trajectories that explicitly reference the image before performing algorithmic computation, and that revise earlier interpretations when contradictions arise. \textbf{These behaviors emerge without any chain-of-thought supervision}. The best-performing reward configuration improves accuracy on Qwen-2.5-VL-7B by 5.56 absolute points over the base model. These gains, however, do not generalize uniformly. First, reward variants that include a verifiable accuracy signal improve in-domain accuracy at a measurable cost to cross-domain transfer, suggesting overfitting to the training distribution rather than general visual reasoning improvement. Second, on categories where perceptual difficulty is highest (e.g., \textit{Clock}), no reward variant produces substantial gains. Qualitatively, we show that the model often produces long, internally consistent reasoning trajectories that proceed from an initial misreading of the image. In other words, though the structure of the reasoning is correct, its \textit{inputs} are not. These limitations suggest that reward signals applied at the level of the answer or the reasoning chain do not reliably correct errors that originate in the perceptual stage. As such, further progress in this setting will likely require reward signals that evaluate the fidelity of the model's visual interpretation directly. Overall, this work contributes:

\begin{itemize}

    \item \textbf{Empirical evidence} that \textbf{isolates the contribution of visual perception} to multimodal puzzle-solving by substituting images with deterministic textual renderings that preserve task-relevant information. This substitution yields average gains of over 20 points for Claude models across nine puzzle categories, indicating that perception, rather than algorithmic reasoning, is the dominant constraint on performance.
    
    \item \textbf{Design and evaluation of six reward variants} that probe distinct hypotheses about how reward signals interact with visual grounding, and demonstrate that reward design alone (without chain-of-thought supervision) induces extended, image-grounded reasoning trajectories and improves Qwen-2.5-VL-7B by 5.56 absolute points.
    
    \item An \textbf{analysis of systematic limitations} of this approach, showing that reward variants with verifiable accuracy signals trade in-domain gain for cross-domain transfer, and reasoning trajectories frequently inherit rather than correct upstream perceptual errors. These results suggest that reward signals targeted at perceptual fidelity are needed for further progress.
\end{itemize}

\section{Design of Reward Functions}
\label{sec:reward-functions}


We design six reward functions that vary along three complementary axes: (i) how the final answer is scored: by exact match, by partial credit for near-correct answers, or not at all; (ii) what intermediate output structure is rewarded: a single \texttt{<think>} block, or multiple stages spanning image description, reasoning, and self-reflection; and (iii) whether visual grounding is explicitly rewarded through a dedicated \texttt{<visual>} tag. Each variant thus probes a distinct hypothesis about which components of the reward signal contribute to long, image-grounded multimodal reasoning. Table~\ref{table:reward_summary} summarizes the six variants and their motivating questions; all are implemented within a single modular scoring framework, with prompts and pseudo-code provided in Tables~\ref{table:prompts} and~\ref{table:code-sample}.

\subsection{Reward Variants and Implementation}

\paragraph{\rFormat:} Following the formulation in \citet{deepseekai2025deepseekr1incentivizingreasoningcapability}, this variant rewards both the presence of a \texttt{<think>} block and exact-match accuracy on the \texttt{<answer>} content. It tests whether the standard text-domain recipe (i.e. format plus accuracy) is sufficient in the multimodal setting.

\paragraph{\rOutcome:} This variant rewards only exact-match accuracy on the \texttt{<answer>} content, with no credit for intermediate structure. It isolates the contribution of the accuracy signal alone and tests whether reasoning structure can be elicited from the prompt without being explicitly rewarded, motivated by DAPO~\cite{yu2025dapoopensourcellmreinforcement}.


\paragraph{\rStructured:}  This variant distributes credit across three
reasoning stages (\texttt{<image description>}, \texttt{<think>},
and \texttt{<rethink>}) in addition to the final answer. Each stage
is scored for both multiple-tag presence and the amount of unique
content within the tag, with \textbf{repetition penalized and counts smoothed
by $\tanh(\cdot)$ to prevent reward hacking through redundant output}. Answer correctness is assigned the highest weight in the composite to maintain factual accuracy. The goal is to test whether explicitly rewarding multi-stage structure induces longer, more grounded reasoning trajectories without chain-of-thought supervision.


\noindent

\paragraph{\rGraded:} Identical to \rStructured but replaces exact-match accuracy with partial credit for near-correct answers (e.g. prediction of 4 or 6 when ground truth is 5). All other components (repetition penalties, $\tanh(\cdot)$ smoothing, and stage-wise structural rewards) remain unchanged, to test whether softer accuracy supervision improves reasoning, e.g. on tasks with dense numeric answer spaces.


\paragraph{\rGrounded:} This reward focuses on visual grounding by requiring models to integrate image-specific information into their reasoning. It introduces an explicit \texttt{<visual>} tag inside the \texttt{<think>} block and rewards unique visual descriptions within it, alongside the standard \texttt{<think>} and exact-match accuracy rewards. The same repetition penalty and $\tanh(\cdot)$ smoothing from \rStructured also apply. This tests whether explicitly rewarding image-referent content within the reasoning chain improves visual grounding.


\paragraph{\rProcess:} Finally, this configuration removes the accuracy signal entirely. The model is rewarded only for unique content in the \texttt{<image description>}, \texttt{<think>}, and \texttt{<rethink>} tags, and receives full credit for producing any \texttt{<answer>} tag regardless of its content. This effectively tests a lower bound on supervision: whether reasoning behavior can emerge from structural rewards alone, without any verifiable signal on correctness.

\section{Dataset Collection and Generation}
\label{section:dataset-generation}
We use the \texttt{AlgoPuzzleVQA} dataset introduced by \citet{ghosal-etal-2025-algopuzzlevqa}, which comprises challenging visual puzzles designed to evaluate the capabilities of MLLMs in visual perception, language comprehension, and algorithmic reasoning. For our experiments, we selected nine categories from \texttt{AlgoPuzzleVQA} based on their suitability for conversion into text-only representations. This enables us to isolate and examine the reasoning bottlenecks of MLLMs without relying on visual inputs. The selected categories include: Clock, Checker Move, Maze, Move Box, N-Queens, Number Slide, Tower of Hanoi, Rotting Kiwi, and Water Jugs. Descriptions and examples are provided in Appendix \ref{appendix:dataset-details}.

\paragraph{Training Data Generation.}
The dataset we use (\texttt{AlgoPuzzleVQA}) provides only 100 visual samples per category, which is insufficient for RL training. To scale up the dataset, we generated additional training samples by adapting the source code\footnote{https://tinyurl.com/4e5cvevv} with necessary modifications. Specifically, we created 9,000 new training samples across four categories: Clock, Maze, Move Box, and N-Queens. The Clock category contains 3,000 samples, while each of the other three categories includes 2,000 samples. These samples were generated by varying features such as the clock's shape, the color and size of the clock hands, the orientation of the grid, and the positions of the boxes and queens. The original AlgoPuzzleVQA dataset is retained as the gold test set, and we ensured that none of the generated training samples overlap with the test data. Additionally, all training examples are unique and contain no repetitions. 
\section{Experimental Setup}
We design experiments to address three questions. First, to what extent are the limitations of MLLMs due to challenges in visual perception or algorithmic reasoning? Second, can different reward functions affect RL training to unlock long visual reasoning? Third, how well does the RL-trained model generalize to cross-distribution classes?

As our \textbf{baseline}, we evaluate the \texttt{Qwen-2.5-VL-7B} model using Chain-of-Thought prompting (\textit{<image description>, <think>, <rethink>, <answer>}) without any RL fine-tuning. For the \textbf{reward-function analysis}, we apply GRPO directly to the base model without supervised fine-tuning (SFT) or a cold-start stage, as no SFT data is available. We discuss the details in Section \ref{training-settings}. 

Finally, we extract the answer from the model's prediction and compare it with the ground truth using exact match, reporting Accuracy as the evaluation metric \cite{toh2025jumpingreasoningcurvetracking}. Prior work by \citet{ghosal-etal-2025-algopuzzlevqa} evaluated model performance in both multiple-choice (MCQ) and open-ended settings. In our experiments, we found that in the MCQ setting, models often guessed the correct answer from the provided options, even when their reasoning was incorrect. This limits the evaluation’s ability to accurately reflect the model’s true capabilities. Therefore, we evaluate all models exclusively in the open-ended setting, where no answer options are provided. We recommend that future work adopt the open-ended format to accurately assess models' performance.

\subsection{Bottleneck Experiments}
\begin{figure}[t!]
  \centering
  \includegraphics[width=1.0\linewidth]{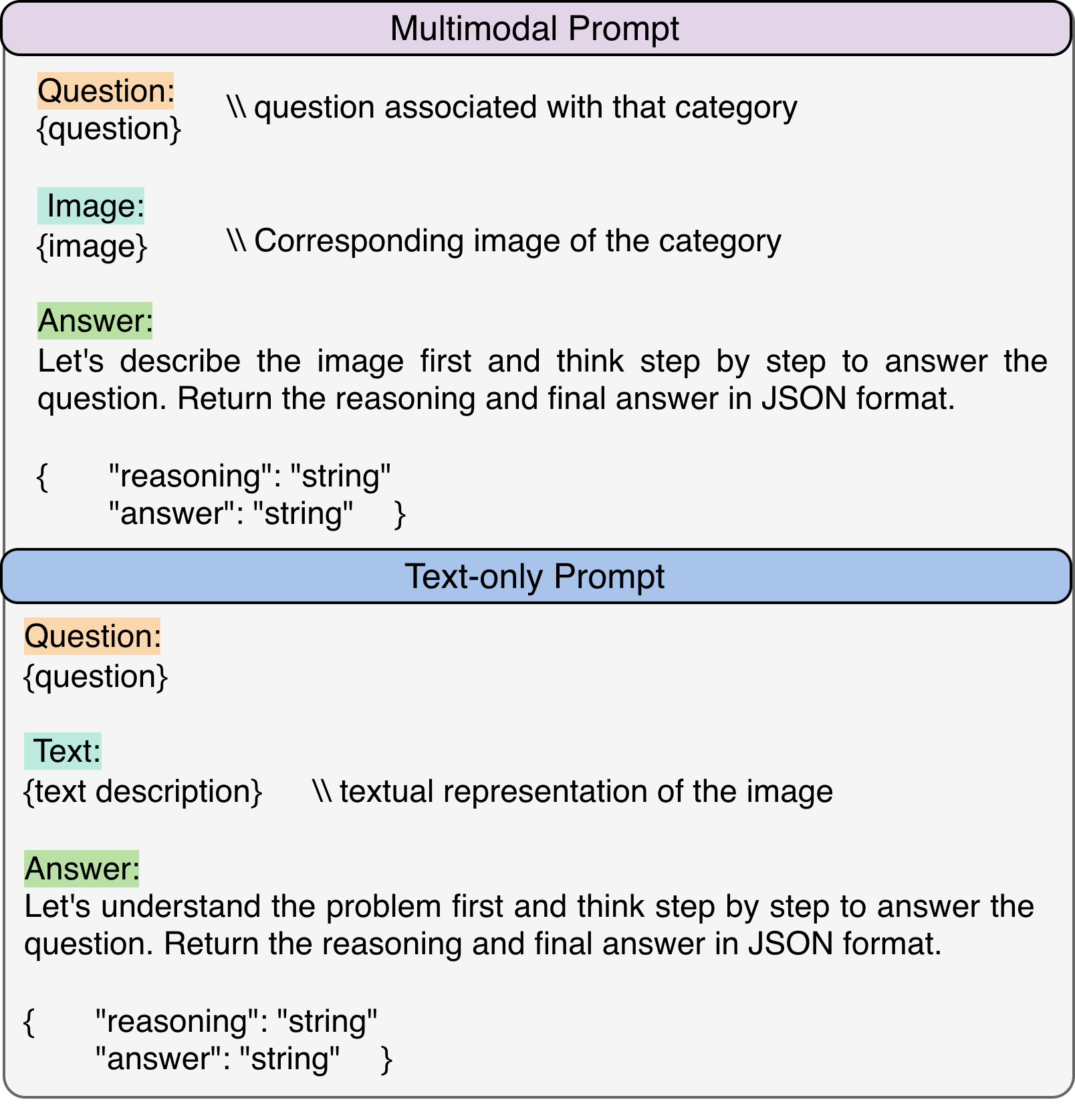}
 \caption{Generic prompt structure used across all categories in both multimodal and text-only settings. 
 }
 \label{figure:generic-prompt}
\end{figure}

To analyze the reasoning bottlenecks, we evaluate \texttt{Claude-3.5-Sonnet} and \texttt{Claude-3.7-Sonnet} in \texttt{thinking} mode across the nine selected categories. We conduct the experiments in two settings: i) \texttt{Multimodal} (\texttt{<image, question>}): where the image and corresponding question are provided to the model, ii) \texttt{Text-only} (\texttt{<text-representing image, question>}): where the image is converted into a text description, which is then passed to the model along with the question. For instance, in the `Checker-Move' category, text descriptions are simple strings where colors are encoded as characters—`R' for red, `G' for green, and `-' for white space. Other categories follow similarly intuitive character-based representations. Examples of how images are converted into character-based textual representations are provided in Table~\ref{table:image-text-conversion}. Generic prompt structure is illustrated in Figure~\ref{figure:generic-prompt}.

\subsection{Training Settings}
\label{training-settings}
We use the \texttt{Qwen-2.5-VL-7B} (instruct) model for our experiments~\cite{qwen2.5}. We train the model using the GRPO \cite{deepseekai2025deepseekr1incentivizingreasoningcapability}, a reinforcement learning–based optimization method, with synthetically generated training samples (Section~\ref{section:dataset-generation}). Building on this setup, we examine how reinforcement learning may influence the model’s reasoning behavior by constructing two datasets, each aligned with a complementary hypothesis. First, we hypothesize that training on challenging, high-difficulty samples can help unlock the model's longer reasoning abilities. Since Clock is one of the categories where the base model struggles most, we construct a training set consisting of 3,000 samples exclusively from this category, referred to as \textbf{Clock-3k}. Second, we hypothesize that exposure to a diverse set of reasoning categories during RL training could encourage the model to make more generalizable and transferable improvements in reasoning. To test this, we create a second training set, \textbf{Diverse-8k}, comprising 8,000 samples distributed evenly across four categories: Clock, Maze, Move-Box, and N-Queens. Further details on hyperparameters are provided in Appendix~\ref{appendix:implementation-details}.


\begin{table}[!htb]
\centering
\renewcommand*{\arraystretch}{0.9}
\setlength{\tabcolsep}{3pt}
\small
\begin{tabular}{l|R{1cm}R{1cm}R{1cm}R{1cm}}
\toprule
& \multicolumn{2}{c}{Claude-3.5} & \multicolumn{2}{c}{Claude-3.7 (thinking)}\\
\midrule
Category& MM & TO & MM & TO \\
\midrule
Clock & 3 & 84 & 7 & 83 \\ 
Checker move & 16 & 9 & 39 & 44 \\
Maze &10 & 15 & 14 & 26 \\
Move box &33 & 58 & 34 & 70 \\
N-queens & 25 & 95 & 25 & 97 \\
Number slide & 36 & 60 & 74 & 77 \\
Tower of Hanoi & 44 & 60 & 85 & 85 \\
Rotting kiwi & 19 & 37 & 32 & 50 \\
Water Jugs & 47 & 55 & 72 & 62 \\
\midrule
Avg. & \textbf{25.9} & \textbf{52.6} & \textbf{42.4} & \textbf{66.0} \\
\bottomrule
\end{tabular}
\caption{Comparison of accuracy in multimodal (MM) and text-only (TO) settings in different categories.} 
\label{table:mllm-bottlneck}
\end{table}

\begin{table*}[!h]
\centering
\renewcommand*{\arraystretch}{0.92}
\setlength{\tabcolsep}{5pt}
\small
\begin{tabular}{l|r|rrrrrr|rrrrrr}
\toprule
Category& & \multicolumn{6}{c}{\textbf{Clock-3K Data}} & \multicolumn{6}{c}{\textbf{Diverse-8k Data}}\\
\midrule
 & Base & $\mathcal{R}_F$ & $\mathcal{R}_O$ & $\mathcal{R}_S$ & $\mathcal{R}_G$ & $\mathcal{R}_V$ & $\mathcal{R}_P$ &  $\mathcal{R}_F$ & $\mathcal{R}_O$ & $\mathcal{R}_S$ & $\mathcal{R}_G$ & $\mathcal{R}_V$ & $\mathcal{R}_P$\\
\midrule
Clock         & 3  & \cellcolor{green!50} 37 & 8  & \cellcolor{green!15}13 & 10 & 4  & 1  & 6  & 6 & 4  & 1  & 2  & 3  \\
Checker move  & 6  & 1  & 3  & 4  & 5  & 5  & \cellcolor{green!15}10 &  4 & \cellcolor{green!15}10 & 2 & \cellcolor{green!15}10 & 6 & 1  \\
Maze          & 6  & 0  & 5  & 10 & 7  & 0  & 11 &  \cellcolor{green!15} 26 & 10 & 12 & 9 & \cellcolor{green!50} 30 & 9 \\
Move box      & 25 & 13 & 24 & 21 & 17 & 17 & 19 &  30 & \cellcolor{green!15} 43 & \cellcolor{green!50} 54 & 30 & 20 & 21 \\
N-queens      & 8  & 2  & 6  & 7  & 6  & 6  & 5  &   \cellcolor{green!50} 17 & 11 & 15 & \cellcolor{green!15} 16 & 13 & 9 \\
Number slide  & 13 & 4  & \cellcolor{green!15} 14 & \cellcolor{green!50} 16 & 10 & 10 & 11 &  7 & 10 & 10 & 6 & 10 & 9  \\
Tower of hanoi& 1  & 0  & 0  & 0  & 2  & \cellcolor{green!15} 4  & \cellcolor{green!15} 4  &   3 & 2 & 0 & \cellcolor{green!50} 6 & 2 & 0 \\
Rotting kiwi  & 24 & 14 & 22 & 22 & 18 & 20 & 28 &  22 & \cellcolor{green!50} 30 & 27 & 27 & 27 & \cellcolor{green!15} 29 \\
Water jugs    & 3  & 0  & 1  & 17 & 16 & 11 & 13 &   4 & 14 & 16 & \cellcolor{green!50} 27 & 10 & \cellcolor{green!15} 20 \\
\midrule
Avg.       & \textbf{10} & \textbf{7.89} & \textbf{9.22} & \textbf{12.22} & \textbf{10.11} & \textbf{8.56} & \textbf{11.33} & \textbf{13.22} & \textbf{\cellcolor{green!15} 15.11} & \textbf{\cellcolor{green!50}15.56} & \textbf{14.67} & \textbf{13.33} & \textbf{11.22}  \\

\bottomrule
\end{tabular}
\caption{Evaluation results for RL trained Qwen-2.5-VL-7B model across different reward functions. $\mathcal{R}_F$, $\mathcal{R}_O$, $\mathcal{R}_S$, $\mathcal{R}_G$, $\mathcal{R}_V$, and $\mathcal{R}_P$ denote \rewardSet~respectively. The base score indicate the performance with Chain-of-Thought approach. For each row, the best value is highlighted in a dark shade and the second-best in a light shade. All tied values within a class are highlighted in a light shade.
}
\label{table:reward-function-results}
\end{table*}

\section{Results}
\subsection{Perception as primary bottleneck}
Table~\ref{table:mllm-bottlneck} reports the performance (accuracy) of \texttt{Claude-3.5} and \texttt{Claude-3.7} across the nine puzzle categories in both the multimodal (MM) and text-only (TO) settings. Substituting the image with its textual rendering yields large and consistent gains: on average 26.7 points for \texttt{Claude-3.5} and 23.6 for \texttt{Claude-3.7}, The effect is most pronounced on categories with high perceptual demand. On \textit{Clock}, \texttt{Claude-3.7} improves from 7\% to 83\%; on \textit{N-Queens}, both models move to near-perfect accuracy. Since the question and the underlying algorithm are identical across the two conditions, the only factor that has changed is the difficulty of extracting the relevant information from the visual input. This empirically supports the hypothesis that visual perception, rather than algorithmic reasoning, is the dominant constraint.



\paragraph{Qualitative Analysis.}
To corroborate this, we inspected the multimodal-condition trajectories qualitatively, and found further support for this interpretation. Errors most often originate at the perception stage and then propagate through reasoning chains that are otherwise internally valid. For example, in the \textit{Clock} category, models often fail to read the time from the positions of the clock hands; in \textit{N-Queens}, they struggle to correctly identify the queens' positions on the board; and in \textit{Move Box}, they fail to recognize the spatial positions of the person, box, and target flags. Table~\ref{table:bottleneck-errors} presents a few representative examples where failures in visual reasoning lead to incorrect model predictions. Together, these patterns indicate that the reasoning components of these models are largely intact for the algorithmic tasks studied here; the bottleneck is in feeding them correct inputs.



\subsection{Diverse training data yields broad gains}
As illustrated in Table~\ref{table:reward-function-results}, training with the diverse dataset yields improvements across all reward functions compared to training solely on hard samples (Clock-3k). The \rStructured reward achieves the highest average score (15.56), followed closely by the \rOutcome reward (15.11). Interestingly, the \rProcess reward produces unexpected behavior: in both settings, it surpasses the baseline, and in the Clock-3k setting, it outperforms several other reward functions. These findings are consistent with \citet{shao2025spuriousrewardsrethinkingtraining}, which reports that RL training without a verifiable reward signal can still improve performance. Nevertheless, certain categories remain challenging: the model consistently struggles with Clock and Tower of Hanoi tasks, in line with \citet{fu2025multimodallargelanguagemodels}, indicating that these puzzles are intrinsically difficult due to the demands of abstract temporal and recursive reasoning. Appendix \ref{additional-results} presents additional analysis.

\subsection{Structured rewards elicit visually grounded reasoning, but long reasoning alone does not guarantee accuracy}
Our reward function design incentivizes the emergence of long visual reasoning in the model. Under \rFormat and \rProcess rewards, the model produces minimal intermediate reasoning, often jumping directly to the answer with little explanation of the image. Both \rStructured and \rGraded achieve the intended behavior: as shown in Figure~\ref{completions}, the model generates structured reasoning steps that reference and explain different parts of the image, exhibiting the `think with image' phenomenon~\cite{zheng2025deepeyesincentivizingthinkingimages,fan2025gritteachingmllmsthink}. This behavior emerges across both in-domain and cross-domain categories, indicating that the reward design and training strategy successfully unlock MLLMs' long reasoning capability. However, \rOutcome's accuracy is on par with or better than \rStructured in about half of the classes, especially in croos-domain settings, suggesting that long reasoning alone does not guarantee improved accuracy for MLLMs~\cite{shao2025spuriousrewardsrethinkingtraining}. 



\subsection{An in-domain/cross-domain trade-off}

\begin{table}[b!]
\centering
\renewcommand*{\arraystretch}{0.95}
\setlength{\tabcolsep}{5pt}
\small
\begin{tabular}{l|R{1cm}R{1cm}|R{1cm}R{1cm}}

& \multicolumn{2}{c}{\textbf{Clock-3K}} & \multicolumn{2}{c}{\textbf{Diverse-8k}}\\
\midrule
 &  ID & CD & ID & CD\\
\midrule
Baseline & 3 & 10.75 & 10.5 & 9.4\\
\midrule
\rFormat  & 37 & 4.25 &  19.75 & 8 \\
\rOutcome  & 8 & 9.37  & 17.5 & 13.2 \\
\rStructured  & 13 & 12.12 & 21.25 & 11 \\
\rGraded & 10 & 10.12 & 14.00 & 15.20 \\
\rGrounded & 4 & 9.12 &  16.25 & 11 \\
\rProcess  & 1 & 12.62 & 10.50 & 11.80 \\
\bottomrule
\end{tabular}
\caption{In-domain (ID) and cross-domain (CD) performance of the RL-trained Qwen-2.5-VL-7B model under different reward functions. For Clock-3K, the Clock category is treated as in-domain, while the remaining eight categories are considered cross-domain. For Diverse-8K, the aggregate performance over four categories—Clock, Maze, Move-Box, and N-Queens is reported as in-domain, while the remaining five categories are considered cross-domain.} 
\label{table:in-ood-performance}
\end{table}




To evaluate generalization, we measure performance on both in-domain (seen) and cross-domain (unseen) classes. As shown in Table~\ref{table:in-ood-performance}, training on a single-category dataset (Clock-3k) results in limited generalization: although \rFormat achieves a large in-domain gain on the trained category (3 $\to$ 37 on \textit{Clock}), its cross-domain accuracy drops well below the no-RL baseline (10.75 $\to$ 4.25), and 3/6 variants in total fall below the baseline on cross-domain. In contrast, the Diverse-8k dataset substantially improves performance across nearly all classes, outperforming the baselines in both in-domain (ID) and cross-domain (CD) settings, with 5/6 variants improving CD relative to baseline. The pattern across both datasets reveals a trade-off: variants that achieve the largest in-domain gains tend to generalize less well, while variants with weaker accuracy signals (\rProcess, \rGraded) show stronger or more balanced CD performance. This is consistent with overfitting to the training distribution during RL training, an effect mitigated but not eliminated by training on a diverse set of categories. \textbf{Notably, the model demonstrates emergent reasoning capabilities across both in-domain and cross-domain categories (Figures \ref{reasoning-move-box} and \ref{reasoning-rotting-kiwi}).}



\begin{figure}[t!]
  \centering
  \includegraphics[width =\linewidth]{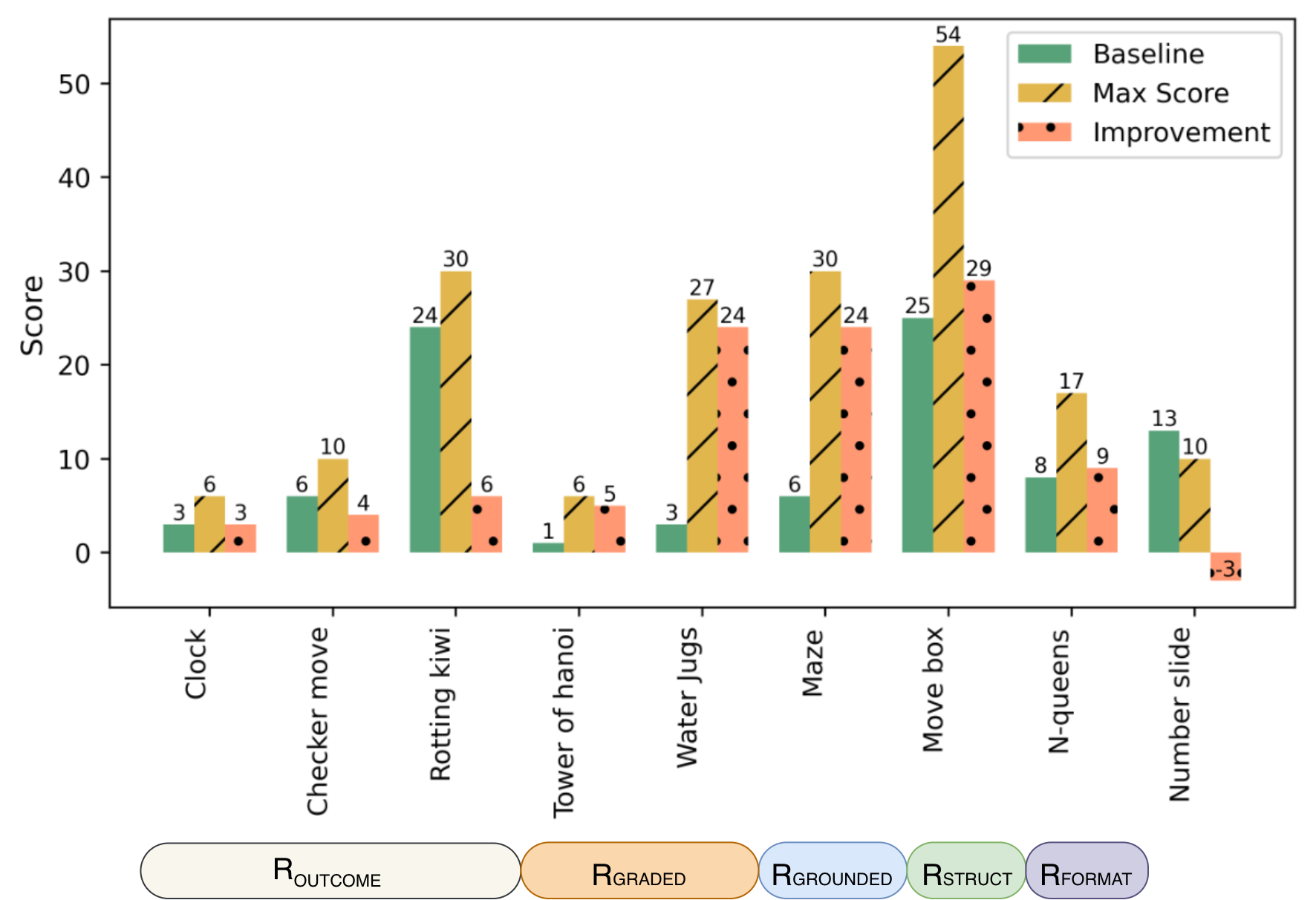}
 \caption{Performance comparison across categories showing baseline vs. maximum scores and improvement margins. Reward functions that achieved the maximum score are indicated below each category. For the number slide category no reward improve over the baseline. Results are reported for models trained on the Diverse-8K dataset.}
 \label{figure:final-class-results-summary}
\end{figure}

To further investigate generalization, we explore the best-performing reward function for each category with the model trained on the Diverse-8K dataset. Detailed class-specific results are provided in Figure~\ref{figure:final-class-results-summary}. No single reward function consistently dominates across all classes; some reward functions excel in certain categories but underperformed in others. Five different reward functions achieved the maximum score in different categories. For the number slide class, performance dropped with RL training. This variability highlights the heterogeneity in task difficulty and model behavior. The variability in reward function performance arises because puzzle categories differ in their reasoning demands. We group them into two types:

\begin{itemize}
    \item \textbf{Visual dependency} (e.g., Move Box, Maze): These puzzles require correctly reading spatial layouts, object positions, and grid configurations from the image. Rewards such as \rStructured or \rGrounded, which encourage visual reasoning, are particularly effective for these tasks because they push the model to attend to relevant image details before reasoning. Table \ref{table:visual-performance} shows the effectiveness on reasoning oriented rewards compare to outcome-only reward and baseline.
    \item \textbf{Algorithmic dependency}: RL training generally improves performance over the baseline across these categories (Table~\ref{table:reward-function-results}). However, these puzzles (e.g., Checker Move, Tower of Hanoi) have relatively simple visual inputs but require more complex recursive or temporal reasoning. In such cases, encouraging additional visual reasoning may provide limited benefit as they might distract the model from the algorithmic steps needed to solve the task.
\end{itemize}
Appendix~\ref{additional-results} presents additional analysis of how rewards affect performance. 

\begin{table}[h!]
\centering
\renewcommand*{\arraystretch}{0.95}
\setlength{\tabcolsep}{5pt}
\small
\begin{tabular}{l|cccC{2cm}}

\midrule
Category & CoT & $\mathcal{R}_O$ & $\max$({$\mathcal{R}_S$, $\mathcal{R}_G$, $\mathcal{R}_V$)}\\
\midrule
Maze & 6 & 10 & 30 \\
Move box & 25 & 43 & 54 \\
Rotting kiwi & 24 & 30 & 27 \\
Water jugs & 3 & 14 & 27 \\
\bottomrule
\end{tabular}
\caption{Head-to-head comparison of the outcome-only reward (\rOutcome) and reasoning-oriented rewards (\rStructured, \rGraded, \rGrounded) on visually dependent categories. The best reasoning-oriented reward improves performance by up to 29 points over the CoT baseline and consistently outperforms the outcome-only reward illustrating the impact of longer reasoning.} 
\label{table:visual-performance}
\end{table}

\subsection{Error Analysis}
 Despite being our best-performing model, the RL-trained Qwen-7B achieves only 15.56\% accuracy. To understand this limited performance, we qualitatively analyze the errors and identify four primary causes of failure. First, the model makes visual understanding errors even after generating extensive visual reasoning, it incorrectly identifies object positions or attributes in the input image. See the generated completion with the \rStructured reward in Figure~\ref{completions}. Second, the model also fails at algorithmic reasoning, generating erroneous intermediate computations or decision steps.
\begin{tcolorbox}[
    colback=gray!5,
    colframe=red!60!black,
    title=\textbf{Miscalculation of time},
    fonttitle=\bfseries\small,
    sharp corners,
    fontupper=\small,
    boxrule=0.8pt
]
....<intermediate reasoning>....\par
Subtract the Time: Start with the hours. \textcolor{red}{Subtracting 2 hours from 1 o'clock results in 9 o'clock.}\par
....<continue reasoning>....\par
\end{tcolorbox}

\noindent
Third, we observe inconsistencies in which the model's final answer contradicts its prior reasoning. 
\begin{tcolorbox}[
    colback=gray!5,
    colframe=red!60!black,
    title=\textbf{Contradiction with own reasoning},
    fonttitle=\bfseries\small,
    fontupper=\small,
    sharp corners,
    boxrule=0.8pt
]
....<intermediate reasoning>....\par
Therefore, the time is confidently determined to be \textbf{12:35}.\par
 \texttt{<answer>} \textcolor{red}{12.37} \texttt{<answer>}
\end{tcolorbox}

\noindent
Finally, for a small number of cases, the model fails to produce a valid answer, continuing to generate tokens without reaching a definitive conclusion. 

\begin{tcolorbox}[
    colback=gray!5,
    colframe=red!60!black,
    title=\textbf{Continue generation without answer},
    fonttitle=\bfseries\small,
    fontupper=\small,
    sharp corners,
    boxrule=0.8pt
]
 ....<intermediate reasoning>....\par
The hour hand is closest to the 1.\par
The hour hand is not aligned with the 11\par
....<continue reasoning>....\par
\textcolor{red}{\texttt{Do not generate any <answer> tag.}}
\end{tcolorbox}

Overall, the error analysis highlights the multifaceted challenges posed by visual puzzle tasks. Our findings suggest that, beyond encouraging extended reasoning, this task may require reward signals that explicitly incentivize accurate visual grounding. For example, evaluating the intermediate reasoning with a strong LLM-based judge. Incorporating such signals represents a promising direction for future research.

\section{Related Work}
Research on the visual reasoning abilities of MLLMs has grown rapidly~\cite{chen2024visual,li2025vocot,zhang2024far}, with benchmarks spanning abstract visual puzzles~\cite{song2025visualpuzzlesdecouplingmultimodalreasoning}, diagrammatic and symbolic reasoning~\cite{li2025visiomath}, and algorithmic visual reasoning~\cite{ghosal-etal-2025-algopuzzlevqa}. Across these settings, MLLMs consistently struggle to uncover the visual relations necessary for correct inference. While CoT prompting elicits reasoning effectively in text-only domains~\cite{wei2022chain, kojima2022large}, its visual extensions often produce interpretations weakly grounded in the image~\cite{dong2025insight, zhang2024far}. Because existing studies evaluate on prediction accuracy alone, they offer limited insight into whether errors stem from visual misinterpretation or faulty algorithmic reasoning~\cite{ghosal-etal-2025-algopuzzlevqa,bi2025diagnosing}. We instead perform a systematic inspection to isolate the source of error.


Reinforcement Learning has recently emerged as a way to improve reasoning in MLLMs. This is primarily motivated by the advances in process supervision and verifiable reward modeling from text-only systems~\cite{lightman2023let,deepseekai2025deepseekr1incentivizingreasoningcapability}. In the visual setting, these ideas are adapted by using reward signals that verify whether intermediate reasoning steps are consistent with the image. For instance, vision-language process reward models~\cite{wang2025visualprm,ong2025training} provide feedback on reasoning steps by checking them against visual evidence. In parallel, some studies explore multimodal extensions of R1-style training to elicit longer reasoning trajectories for diagrams, mathematical problems, and synthetic visual tasks~\cite{yang2025r1,shen2025vlm,zhou2025r1zerosahamomentvisual}. Complementary approaches such as GRIT~\cite{fan2025grit} and DeepEyes~\cite{zheng2025deepeyes} incorporate explicit visual grounding, encouraging models to reference image regions during multi-step explanations. \citet{liang2025modomodomultidomaindatamixtures} optimize the data mixture with verifiable rewards to improve reasoning.   

However, these approaches rely on intermediate supervision such as executable tool outputs, bounding boxes, or annotated reasoning traces. An open question thus remains as to whether structured reasoning can be induced through reward design. We take a step toward addressing this by designing reward functions and systematically evaluating their impact on encouraging long, structured visual reasoning without requiring additional supervision.

\section{Conclusion}
Our results suggest a pressing need to decouple perception and reasoning in post-training for multimodal models. Today's conventional recipes, largely adapted from text, apply reward at the level of the answer or the reasoning chain and leave the perceptual stage to improve as a side effect. We argue this is not sufficient: long, structured reasoning can faithfully propagate whatever the model \textit{thought} it saw. The six reward variants we study begin to establish how supervision can be distributed across these stages and isolate which signals shape which behaviors. We see them as a first step toward a broader program in which perception and reasoning are trained, rewarded, and evaluated as the distinct competencies they are.

\section*{Limitations}

Our primary objective is to investigate whether reward-based RL fine-tuning can encourage models to perform long, multi-step visual reasoning required to solve complex puzzles. While alternative approaches such as SFT could potentially boost performance on a specific dataset, they cannot guide a model to exhibit step-by-step reasoning behavior. Another possibility could have been SFT with explicit reasoning traces. However, for visual puzzle datasets, collecting high-quality chain-of-thought annotations is extremely challenging and costly, mainly because the reasoning is often multi-hop, brittle, and domain-specific. For these reasons, we focus on reward-driven RL fine-tuning, where the aim was not merely to fit the dataset but to induce emergent long-horizon reasoning behavior, which we believe is better aligned with the nature of visual puzzle solving.

Another concern is that the textual representations of images are highly structured and likely overstate the perception gap. However, the purpose of this experiment is to establish an upper bound that quantifies the maximum performance gap attributable to visual perception. The text representations are kept minimal (e.g., ``G-RR'' for checker states, simple grid characters); they encode the same spatial information present in the image without introducing any reasoning shortcuts. Moreover, there is no perfect middle ground between an image and structured text for isolating the visual perception bottleneck from reasoning capability. Possible alternatives, such as noisy or partially corrupted text descriptions or caption-based representations generated by a vision model, each introduce their own biases and assumptions, making it difficult to draw clean conclusions. Our approach instead provides a clear and interpretable upper bound that cleanly separates the perception bottleneck from reasoning capability.



    

\bibliography{acl_latex}

\begin{figure*}[t!]
  \centering
  \includegraphics[width =0.98\linewidth]{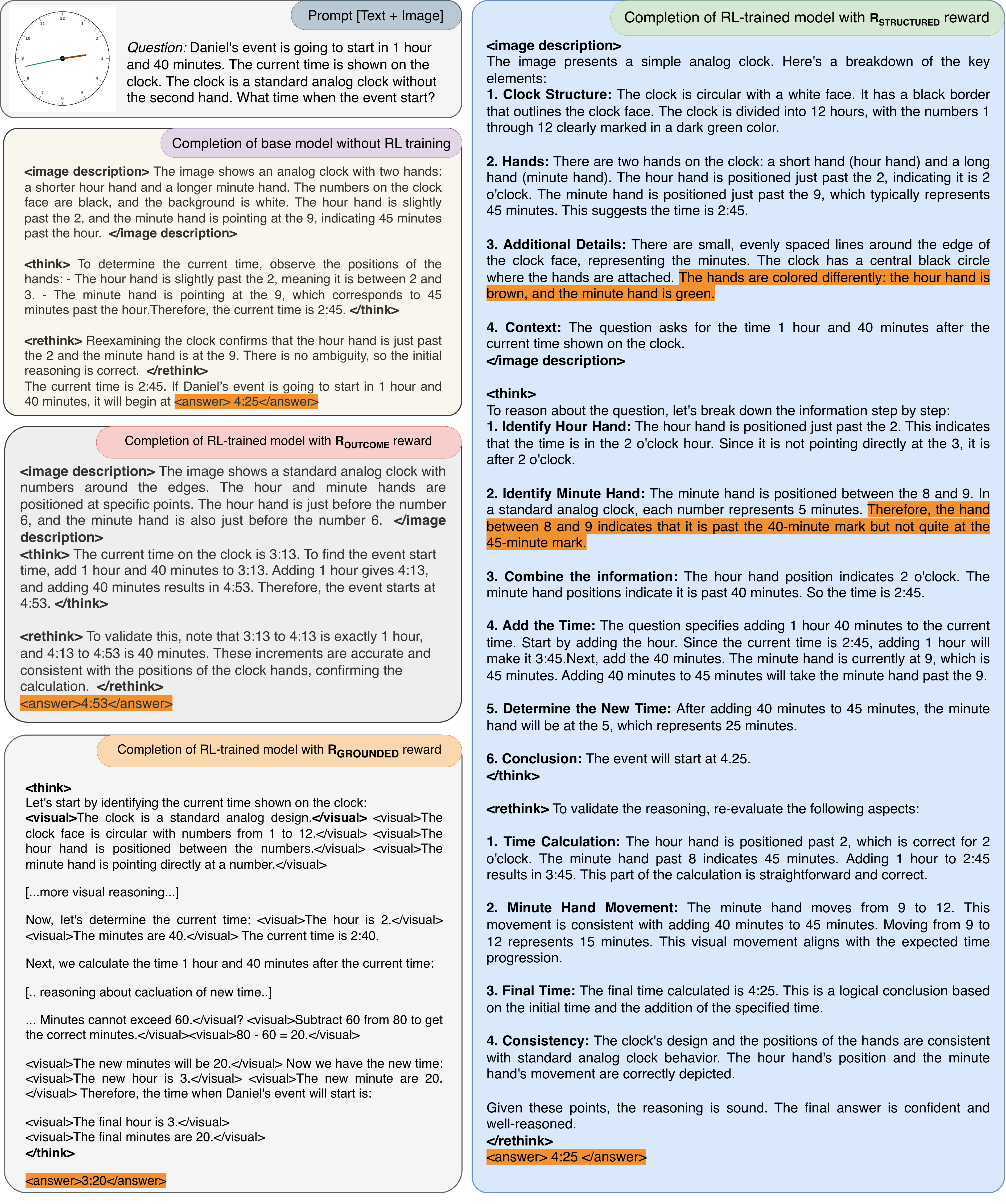}
 \caption{ An example comparing reasoning between the base (Qwen-2.5-VL-7B) with chain-of-thought approach and RL-trained versions with different reward strategies. With the \rStructured reward, the model generates grounded, multi-step visual reasoning--accurately describing fine-grained image details such as the distinct colors of the hour and minute hands. Notably, this structured long-form reasoning emerges naturally from RL training (see Figures \ref{reasoning-move-box} and \ref{reasoning-rotting-kiwi} for additional examples). The model also performs step-by-step self-evaluation, revisiting and verifying each reasoning step before concluding with its final answer.} 
 \label{completions}
\end{figure*}

\newpage
\appendix

\section*{Appendix}
\label{sec:appendix}
\section{Additional Result Analysis}
\label{additional-results}
We perform additional analysis of how reward design affects performance. Tables \ref{table:reward-function-results} and \ref{table:in-ood-performance} surface several patterns worth examining: \rOutcome, which uses no reasoning signal, is often competitive with more complex rewards, and \rProcess, which does not grade the answer, achieves the strongest cross-domain generalization. We account for these as follows.
\begin{itemize}
    \item \textbf{Why \rOutcome is competitive.} This reward leaves the model free to reason however it wants, which suffices for some tasks and yields competitive accuracy. The difference is qualitative rather than quantitative: \rOutcome does not produce the extended reasoning that \rStructured does (Figure \ref{completions}). Accuracy is close, but reasoning quality differs substantially, and that quality is what we aim to improve. 
    \item \textbf{Why \rProcess generalizes in cross-domain.} A strong accuracy signal pushes the RL-trained model toward task-specific answers, raising in-domain accuracy at the cost of transfer. \rProcess sidesteps this by design: it rewards only general reasoning behavior (describing the image, thinking, rethinking), which carries over to unseen tasks. 

    \item \textbf{Why some rewards show large ID–CD gaps.} Rewards with high accuracy weight ($\alpha_a = 0.80$–$0.90$) reinforce task-specific patterns and widen the gap. Rewards that spread weight across reasoning components (\rProcess) or relax accuracy matching (\rGraded) reward more transferable skills, producing more balanced performance

\end{itemize}

\section{Implementation Details}
\label{appendix:implementation-details}
We experimented with two pre-trained MLLMs \texttt{Qwen-2.5-VL-3B} and \texttt{Qwen-2.5-VL-7B}. The 3B parameter model exhibited poor performance in our preliminary experiments and was therefore excluded from further evaluation. We train the 7B model using the GRPO \cite{deepseekai2025deepseekr1incentivizingreasoningcapability} approach with a learning rate of 1e-6 and a batch size of 128. Parameters for training are presented in Table \ref{table:grpo_parameters}. To prevent memory overflow, we set both the maximum prompt length and response length to 2,048 tokens and train the model for 500 steps. The best intermediate model is saved based on the validation score. The KL divergence coefficient is set to 0.0 to promote diverse reasoning strategies. All experiments are conducted on NVIDIA H100 GPUs. We use the EasyR1 \cite{zheng2025easyr1} framework to implement our RL models. 

\begin{table}[h]
\centering
\begin{tabular}{lc}
\toprule
\textbf{Setting} &  \textbf{Value} \\
\midrule
Training Steps & $500$\\
Batch Size & $128$\\
Max Prompt Length & $2048$\\
Max Response Length & $2048$\\
Max Image Pixels & $262144$\\
Min Image Pixels & $65536$\\
Learning Rate & $1e^{-6}$\\
Temperature & $1.0$\\
KL Coefficient & $0.00$\\

\bottomrule
\end{tabular}
\caption{Parameters for the GRPO training}
\label{table:grpo_parameters}
\end{table} 

\begin{table}[h!]
\centering
\small
\begin{tabular}{L{2.5cm}L{4.5cm}}
\toprule
\textbf{Reward Function} &  \textbf{Parameter Weights} \\
\midrule
\rFormat & $\alpha_t$ = $0.10$, $\alpha_a$ = $0.90$ \\
\midrule
\rOutcome  & $\alpha_a$ = $1.0$ \\
\midrule
\rStructured, \rGraded & $\alpha_i = 0.06$, $\alpha_t$ = $0.06$, $\alpha_r$ = $0.08$,  $\alpha_a$ = $0.80$ \\
\midrule
\rGrounded & $\alpha_t$ = $0.10$, $\alpha_v$ = $0.10$, $\alpha_a$ = $0.80$\\
\midrule
\rProcess & $\alpha_i = 0.30$, $\alpha_t$ = $0.30$, $\alpha_r$ = $0.30$,  $\alpha_a$ = $0.10$ \\
\bottomrule
\end{tabular}
\caption{Weight configurations used across different reward functions. To ensure factual accuracy, a high weight is consistently assigned to answer correctness ($\alpha_a$). The other components are weighted based on empirical observations.}
\label{table:reward_parameters}
\end{table}


\section{Dataset Details} 
\label{appendix:dataset-details}
A description of selected categories from AlgoPuzzleVQA is provided below.

\begin{figure}[h!]
  \centering
  \includegraphics[width =0.6\linewidth]{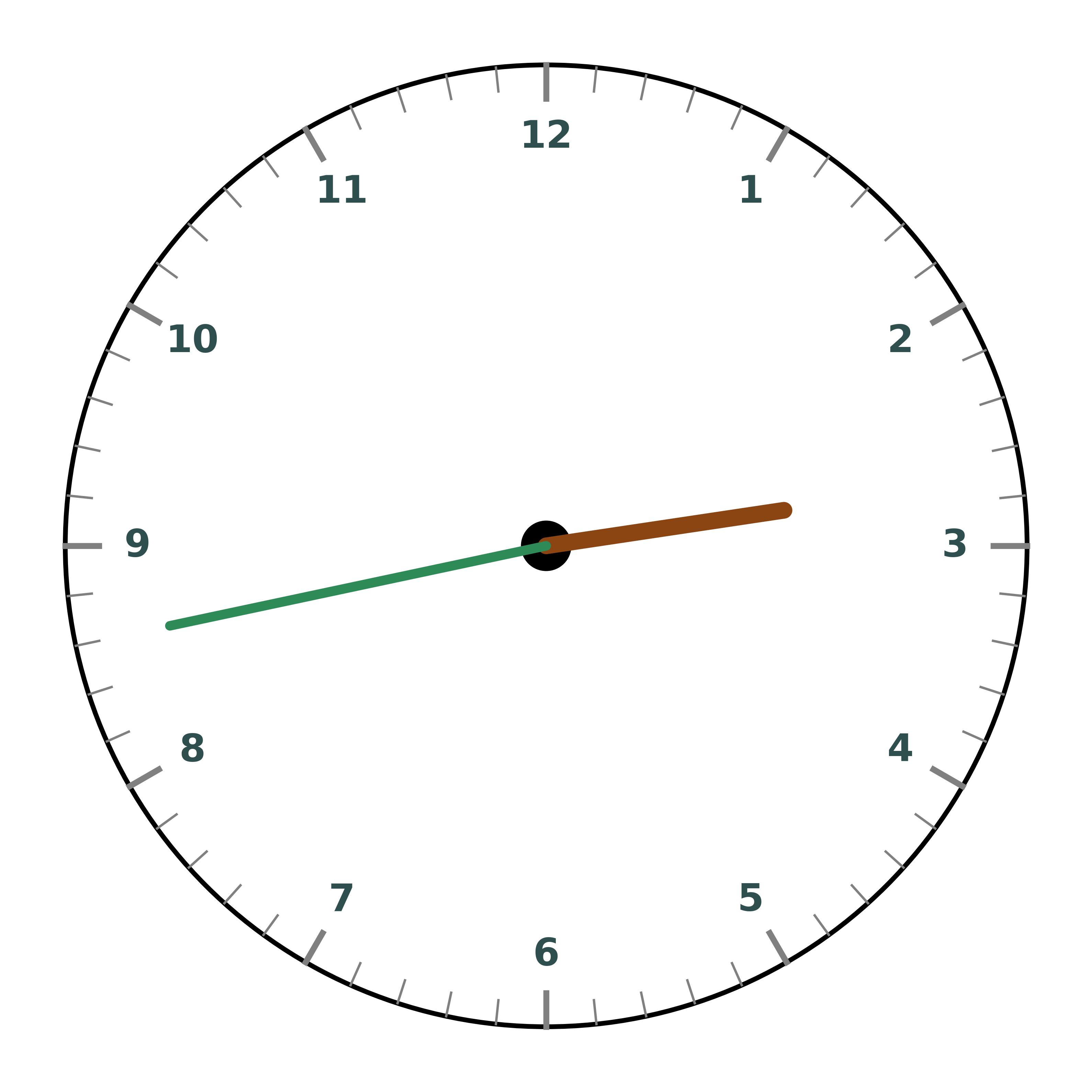}
 \caption{ \texttt{Question: Emily's event is going to start in 1 hour 10 minutes. The current time is shown on the clock. The clock is a standard analog clock without the second hand. What time will the event start?
}}
 \label{clock-example}
\end{figure}

\paragraph{Clock:} This category features analog clocks with only hour and minute hands, displaying a randomly chosen current time. Each question describes an event that occurred (or will occur) \textit{h} hours and \textit{m} minutes before or after the current time, and the model must infer the correct time at which the event took place. An example is shown in Figure \ref{clock-example}.

\begin{figure}[h!]
  \centering
  \includegraphics[width =0.6\linewidth]{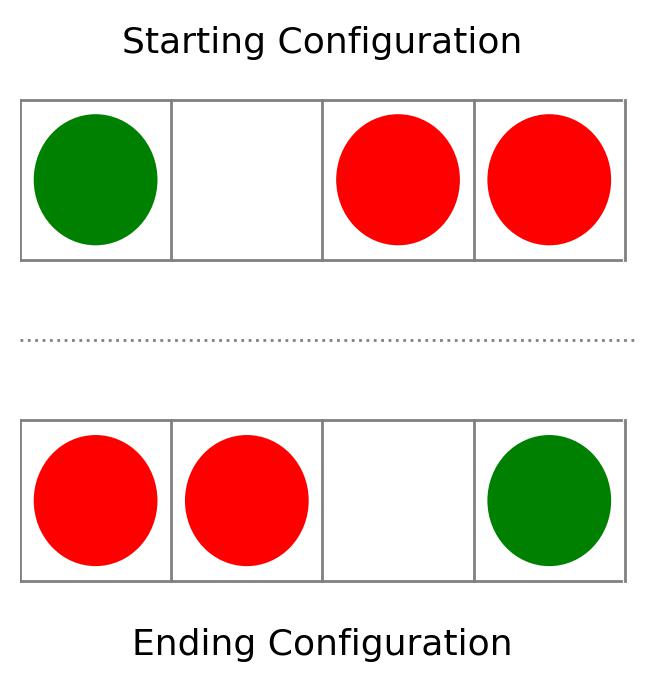}
 \caption{ \texttt{Question: A checker game is being played on a grid of 4 squares with 1 green and 2 red checkers. Initially, the checkers are arranged as shown in the starting configuration, with the 3 checkers occupying 3 squares and one unoccupied square. Green checkers only move rightward and red checkers only move leftward. Every move is either i) a slide to the adjacent empty square, or ii) a jump over one position to an empty square, provided the checker being jumped over is of a different color. Each square can accommodate a maximum of one checker at any time. How many moves are required to reach the ending configuration from the starting configuration following the specified rules?
}}
 \label{check-move-example}
\end{figure}

\paragraph{Checker Move:} This puzzle is played on a one-dimensional grid consisting of \textit{n} cells, where \textit{n–1} cells are initially occupied by either red or green checkers. The objective is to transform the starting configuration into a specified ending configuration, subject to the following constraints: (i) red checkers can only move leftward, and green checkers can only move rightward; (ii) each move must be either a slide into the adjacent empty cell or a jump over one checker of the opposite color into an empty cell; and (iii) each cell may hold at most one checker at any given time. The task is to compute the minimum number of moves required to reach the goal configuration from the initial state. Each puzzle is guaranteed to have at least one valid solution. A sample puzzle is illustrated in Figure~\ref{check-move-example}.

\begin{figure}[h!]
  \centering
  \includegraphics[width =0.7\linewidth]{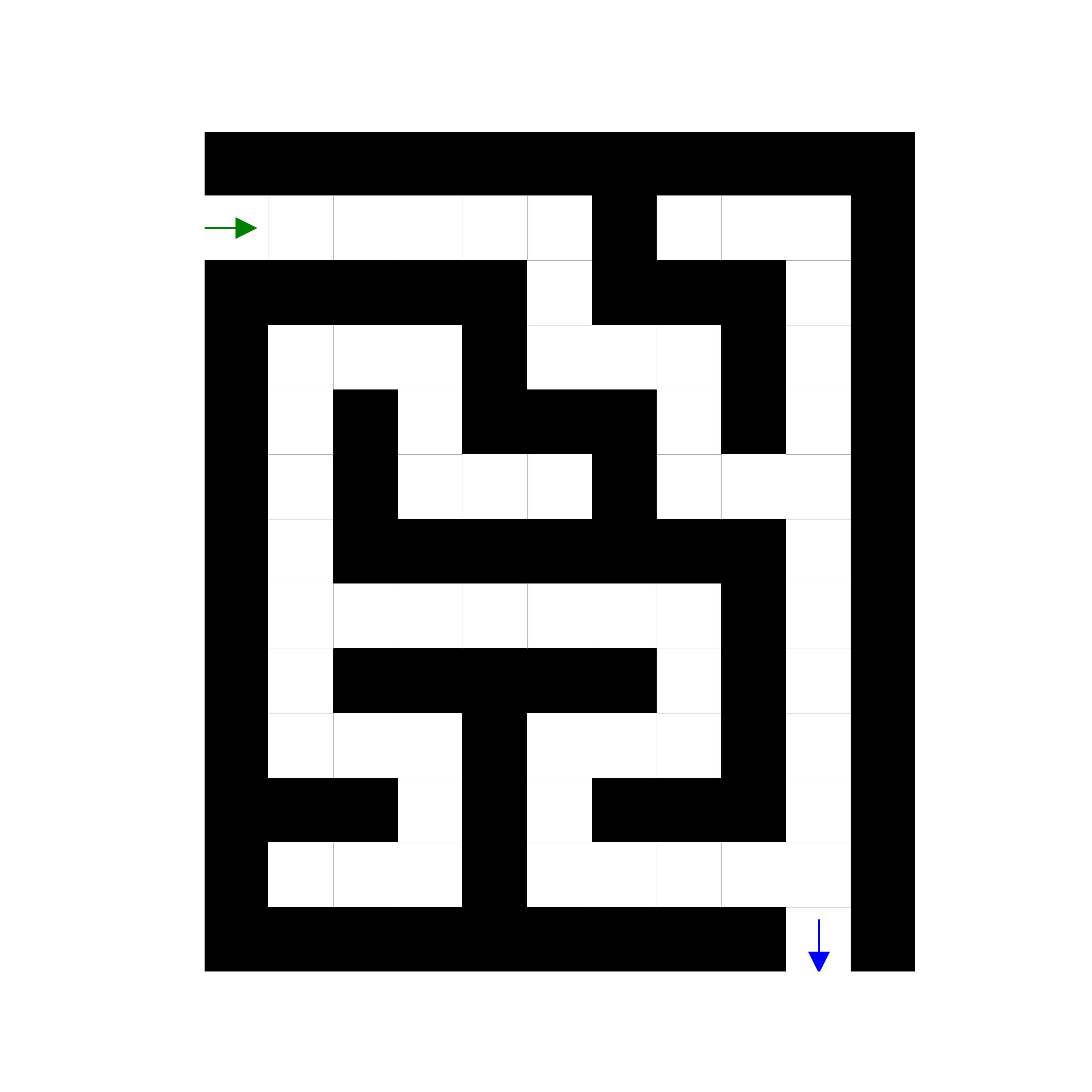}
 \caption{ \texttt{Question: This is maze having 13 * 11 cells. The empty cells are coloured white and the obstacle cells are coloured black. From an empty cell, you can only move up, down, left, or right to another adjacent empty cell. You cannot move diagonally between two empty cells and cannot step into a cell with an obstacle. The entry cell of the maze is shown with the green arrow. The exit cell of the maze is shown with the blue arrow. Suppose you have found the most optimal path in the maze between the entrance and exit, where you need to go through the least number of empty cells and you need to make the least number of left and right turns. What is the total number of left turns do you need to make in this optimal path? }}
 \label{maze-example}
\end{figure}

\paragraph{Maze:} This is a path-finding puzzle with predefined start and end positions. White cells indicate valid paths, while black cells represent walls. The model must answer one of two types of questions: (i) How many left, right, or total turns are required in the optimal path? or (ii) How many cells must be visited in the optimal path, including the entrance and exit? An example maze and corresponding question are shown in Figure~\ref{maze-example}.

\begin{figure}[h!]
  \centering
  \includegraphics[width =0.6\linewidth]{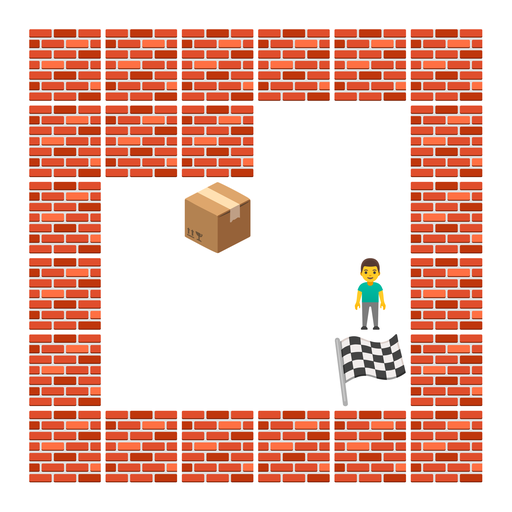}
 \caption{ \texttt{Question: A storekeeper is a puzzle in which the player pushes boxes around in a warehouse trying to get them to target locations. The game is represented by an 6 x 6 grid of characters grid where each element is a wall, floor, or box. Your task is to move the box to the end flag under the following rules: 1. The box can be moved to an adjacent free cell by standing next to the box and then moving in the direction of the box by 1 grid. This is a push. 2. The player cannot walk through the box. What is the minimum number of pushes to move the box to the end flag.}}
 \label{move-box-example}
\end{figure}

\paragraph{Move Box:} This puzzle involves a game with a person, a box, and a target location, all placed on a grid with obstacles. The objective is to move the box to the target location using the minimum number of pushes. This puzzle can be solved using the Breadth-First Search (BFS) technique. An example shown in Figure~\ref{move-box-example}.

\begin{figure}[h!]
  \centering
  \includegraphics[width =0.6\linewidth]{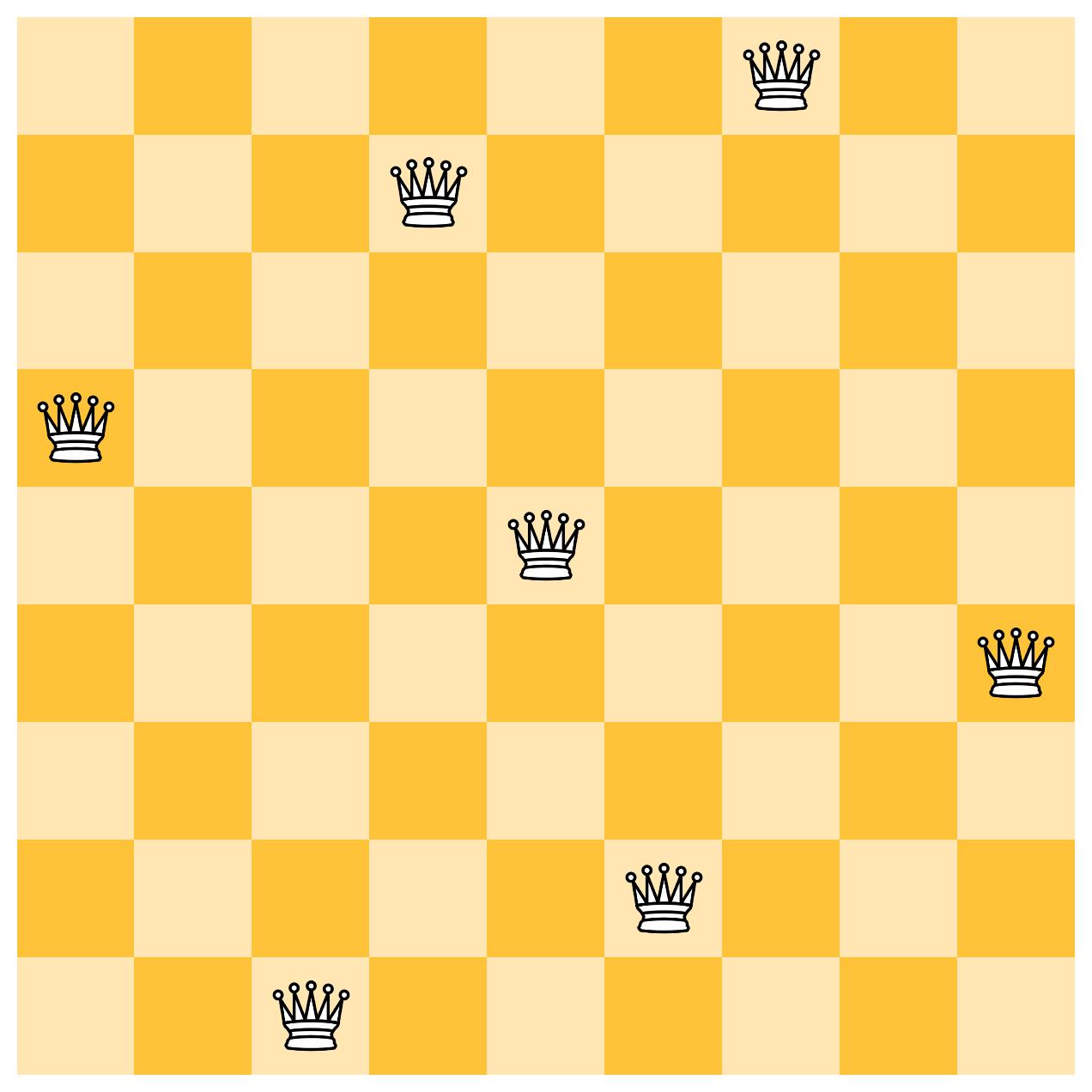}
 \caption{ \texttt{Question: You are given a 9 * 9 chessboard. The Manhattan distance between two squares in a chessboard is equal to the minimal number of orthogonal King moves between these squares on the otherwise empty board. The objective is to place 9 chess queens on this board so that no two queens threaten each other; i.e. no two queens share the same row, column, or diagonal. 7 queens have already been placed in some of the squares of the board, as shown in the image. Suppose you pick two squares to place the two remaining queen pieces in a way that fulfills the objective. What is the Manhattan distance between these two squares?}}
 \label{n-queens-example}
\end{figure}

\paragraph{N-Queens:} This puzzle is based on the classic chess problem where the objective is to place $N$ queens on an $N \times N$ chessboard such that no two queens share the same row, column, or diagonal. In this variation, $N-2$ queens are already placed on the board. The question includes what the Manhattan distance (i.e., the number of unique squares on the board) is between the two remaining queens when they are placed correctly? The solution can be achieved backtracking algorithm. Figure~\ref{n-queens-example} illustrates an n-queens puzzle.

\begin{figure}[h!]
  \centering
  \includegraphics[width =0.6\linewidth]{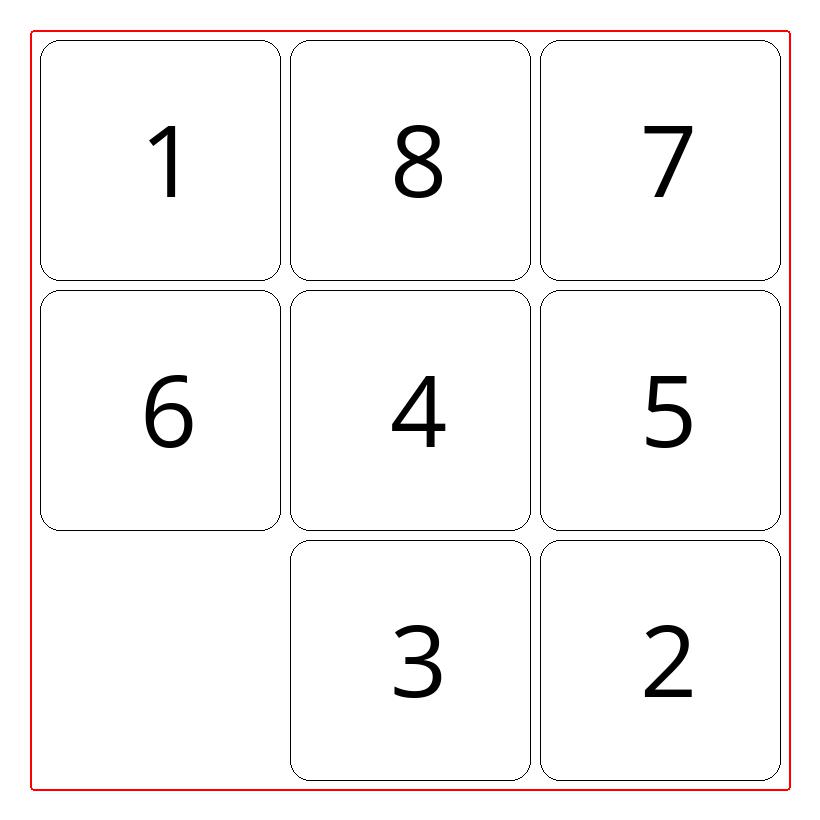}
 \caption{ \texttt{Question: The board shown in the image is a sliding puzzle of 3 * 3 tile dimensions. It has 8 numbered tiles and one unoccupied (open) position. Tiles in the same row or column of the open position can be moved by sliding them horizontally or vertically, respectively. All tiles always stay and move inside the red boundary wall, as shown in the image. A move is defined as moving the open position by one tile unit in any available direction. You start from the board position shown in the image and perform exactly 4 moves such that the open position is seen moving in the following sequence: up, up, right, left. What is the maximum number in the row that now has the open position? }}
 \label{number-slide-example}
\end{figure}

\paragraph{Number Slide:} This puzzle consists of an $N*N$ board with numbered tiles from $1$ to $(N*N) - 1$ and one empty position. The numbers are arranged randomly, and the puzzle involves answering one of the following types of questions: i) How many unique board positions can be reached after performing  $n$ moves? ii) What is the minimum or maximum sum that can be achieved in a particular row or column after performing $n$ moves? iii) Perform $n$ moves where the empty position can be moved in (up, down, left, right) ways. Now, what is the minimum or maximum sum that can be achieved in a particular row or column that has the empty position? Figure~\ref{number-slide-example} shows a $3*3$ puzzle with a sample question.

\begin{figure}[h!]
  \centering
  \includegraphics[width =0.6\linewidth]{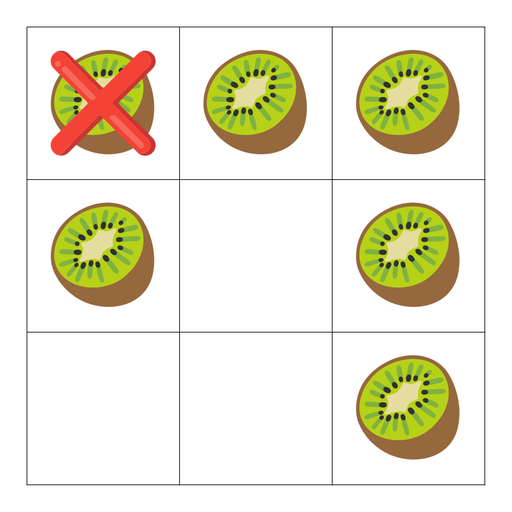}
 \caption{ \texttt{Question: You are given a 3 x 3 grid in which each cell can contain either no kiwi, one fresh kiwi, or one rotten kiwi. Every minute, any fresh kiwi that is 4-directionally adjacent to a rotten kiwi also becomes rotten. What is the minimum number of minutes that must elapse until no cell has a fresh kiwi?}}
 \label{rotting-kiwi-example}
\end{figure}

\paragraph{Rotting Kiwi:}The puzzle involves a rectangular grid containing fresh and rotten kiwis. Initially, one kiwi is rotten. As time progresses, the rot spreads to adjacent kiwis (up, down, left, right) each minute. The objective is to determine the earliest time at which all kiwis become rotten. An example puzzle is illustrated in Figure~\ref{rotting-kiwi-example}.

\begin{figure}[h!]
  \centering
  \includegraphics[width =0.7\linewidth]{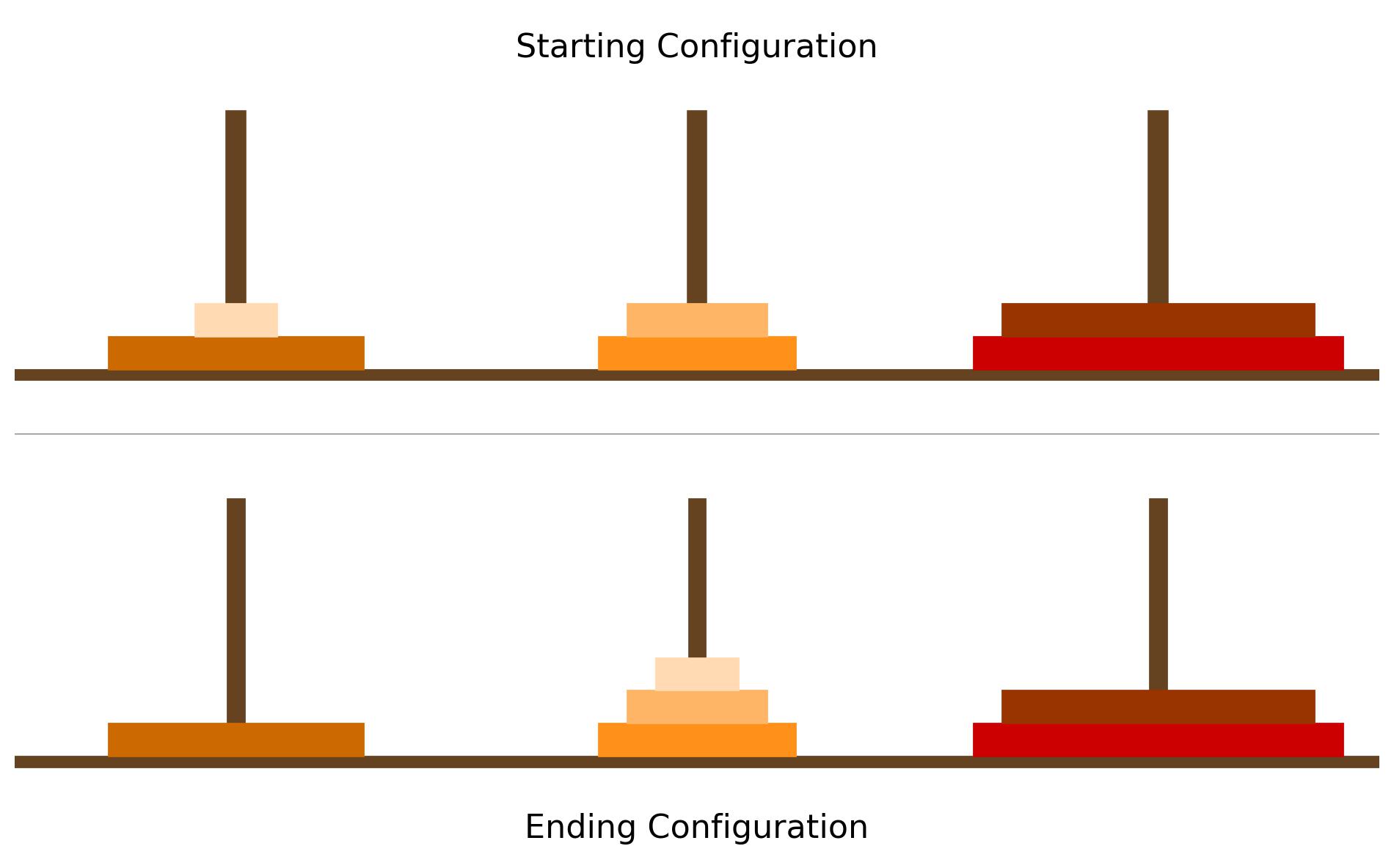}
 \caption{ \texttt{Question: You are playing a Tower of Hanoi game with 3 rods and 6 disks of various diameters, which can slide onto any rod. You are given the starting and ending configuration of the game as shown in the top and the bottom of the image, respectively. The game has the following rules: i) Only one disk may be moved at a time; ii) Each move consists of taking the upper disk from one of the stacks and placing it on top of another stack or on an empty rod; and iii) No disk can be placed on top of a disk that is smaller than it. What is the minimum number of moves required to go from the starting to the ending configuration?}}
 \label{tower-of-hanoi-example}
\end{figure}

\paragraph{Tower of Hanoi:}  This puzzle involves solving a mathematical game consisting of 3 rods and $n$ disks of varying diameters. You are given an initial and a target configuration of the towers. The objective is to reach the target configuration using the minimum number of moves. The game follows these rules: i) Only one disk can be moved at a time, ii) Each move involves taking the top disk from one stack and placing it on top of another stack or an empty rod, iii) No disk may be placed on top of a smaller disk. It is guaranteed that an optimal solution exists. An example puzzle is shown in Figure~\ref{tower-of-hanoi-example}.

\begin{figure}[h!]
  \centering
  \includegraphics[width =0.7\linewidth]{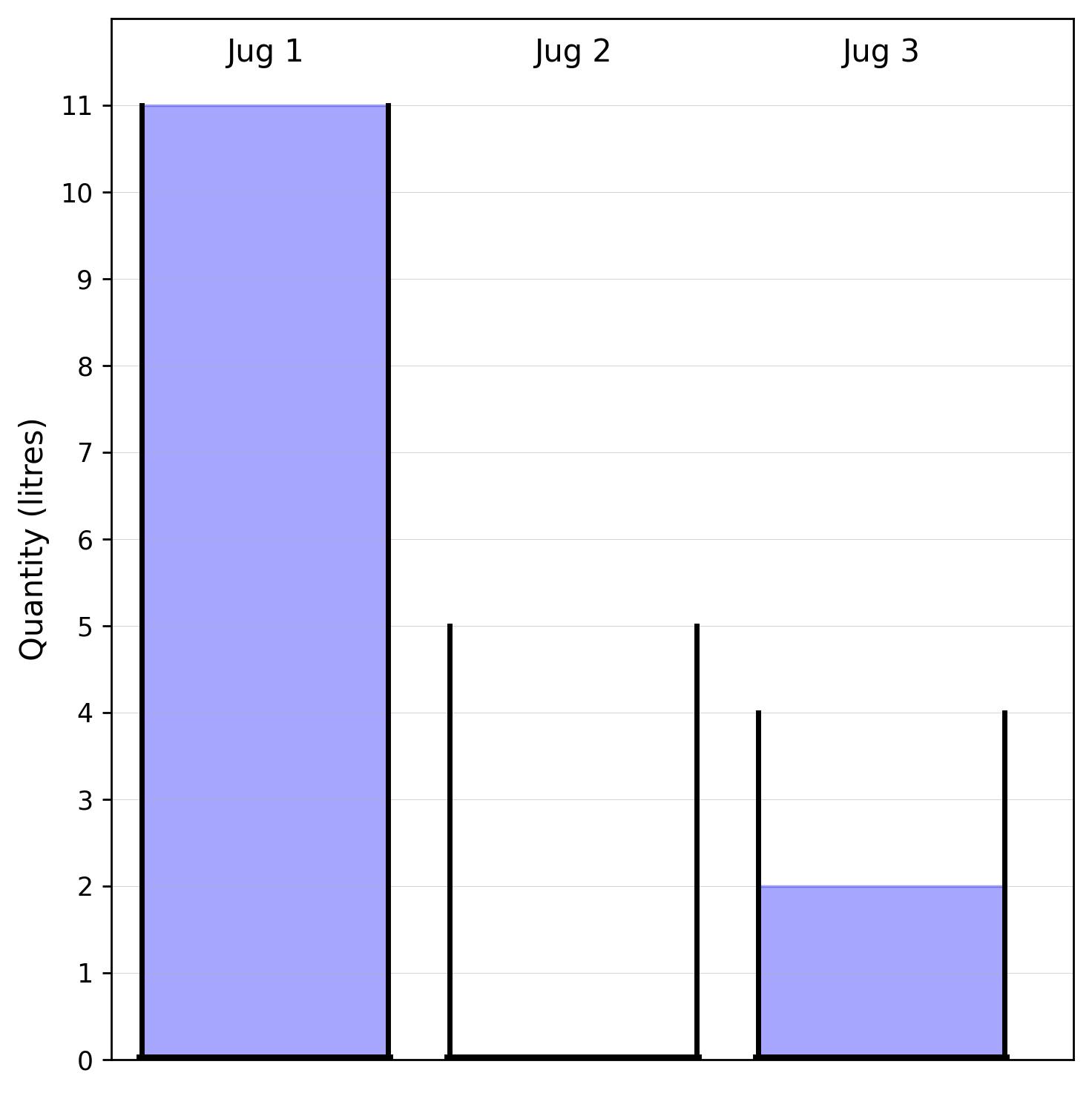}
 \caption{ \texttt{Question: You are given 3 jugs of capacities 11, 5, 4 litres. Initially, the amount of water that is contained in each jar is shown in the image. A single step of water pouring from one jug to another is constrained by the following rules: i) take a non-empty jug and pour water from it to another non-full jug until the first one becomes empty or the second one becomes full; and ii) no water can be split while pouring. The objective is to reach the amounts of 11, 1, 1 litres of water in the jugs from left to right, respectively. What is the minimum number of water pouring steps required to achieve the objective?}}
 \label{water-jugs-example}
\end{figure}

\paragraph{Water Jugs:} This is a measurement-based puzzle involving $n$ jugs with integer capacities. The initial amount of water in each jug is provided as visual context. The model is then asked to determine the number of pouring steps required to reach a goal state, which specifies the desired quantity of water in each jug. Water pouring is subject to the following constraints: i) Water can be poured from a non-empty jug into another non-full jug until either the source jug is empty or the target jug is full, ii) No water can be spilled during the process. Both the initial and goal states are randomly generated with a solution. An example is shown in Figure~\ref{water-jugs-example}.

\clearpage
\begin{table*}[h!]
\centering
\small
\renewcommand{\arraystretch}{0.9}
\begin{tabular}{L{3.7cm}L{3.3cm}L{3.7cm}L{3.2cm}}
\toprule
\textbf{Image} & \textbf{Textual Representation} & \textbf{Image} & \textbf{Textual Representation} \\
\midrule

\includegraphics[width=\linewidth]{figures/clock_0000.jpg} &\texttt{Current Time: 2:43} & \includegraphics[width=\linewidth]{figures/checker_move_0000.jpg} & \texttt{Starting Configuration: G-RR}\newline \texttt{Ending Configuration: RR-G}\\
\midrule
\includegraphics[width=\linewidth]{figures/maze_0000.jpg} &
The maze is illustrated in the following grid, where white empty cells are denoted by \texttt{0} and black obstacle cells are denoted by \texttt{1}.\newline

\texttt{1 1 1 1 1 1 1 1 1 1 1} \newline
\texttt{0 0 0 0 0 0 1 0 0 0 1} \newline
\texttt{1 1 1 1 1 0 1 1 1 0 1} \newline
\texttt{1 0 0 0 1 0 0 0 1 0 1} \newline
\texttt{1 0 1 0 1 1 1 0 1 0 1} \newline
\texttt{1 0 1 0 0 0 1 0 0 0 1} \newline
\texttt{1 0 1 1 1 1 1 1 1 0 1} \newline
\texttt{1 0 0 0 0 0 0 0 1 0 1} \newline
\texttt{1 0 1 1 1 1 1 0 1 0 1} \newline
\texttt{1 0 0 0 1 0 0 0 1 0 1} \newline
\texttt{1 1 1 0 1 0 1 1 1 0 1} \newline
\texttt{1 0 0 0 1 0 0 0 0 0 1} \newline
\texttt{1 1 1 1 1 1 1 1 1 0 1}
 & 
 \includegraphics[width=\linewidth]{figures/move_box_0000.jpg} & 
Grid is shown below, where the wall, floor, and box are represented by characters \texttt{\#}, \texttt{.}, and \texttt{B} respectively. The person is denoted by \texttt{S} and the end flag is by \texttt{T}. \newline

\texttt{\#\#\#\#\#\#} \newline
\texttt{\#\#\#..\#} \newline
\texttt{\#.B..\#} \newline
\texttt{\#...S\#} \newline
\texttt{\#...T\#} \newline
\texttt{\#\#\#\#\#\#}\\

\midrule

\includegraphics[width=\linewidth]{figures/n_queens_0043.jpg} & Queen positions in the chessboard are given in the following lists.\newline
\texttt{[0, 6]} \newline
\texttt{[1, 3]} \newline
\texttt{[3, 0]} \newline
\texttt{[4, 4]} \newline
\texttt{[5, 8]} \newline
\texttt{[7, 5]} \newline
\texttt{[8, 2]}

& \includegraphics[width=\linewidth]{figures/number_slide_0000.jpg} &
The image/board is represented as a grid, and the unoccupied (open) position is denoted with \texttt{0}. \newline
\texttt{1 8 7} \newline
\texttt{6 4 5} \newline
\texttt{0 3 2}\\
\midrule
\includegraphics[width=\linewidth]{figures/rotting_kiwi_0000.jpg} &
Grid is shown below, where \texttt{1} denotes fresh kiwis, \texttt{2} denotes rotten kiwis, and \texttt{0} represents empty spaces.\newline
\texttt{2 1 1} \newline
\texttt{1 0 1} \newline
\texttt{0 0 1}

& \includegraphics[width=\linewidth]{figures/tower_of_hanoi_0000.jpg} & The starting and ending configurations of the towers are given in the following lists.
\newline
\texttt{[4, 1]}, \texttt{[3, 2]}, \texttt{[6, 5]} \newline
\texttt{[4]}, \texttt{[3, 2, 1]}, \texttt{[6, 5]}\\
\midrule

\includegraphics[width=\linewidth]
{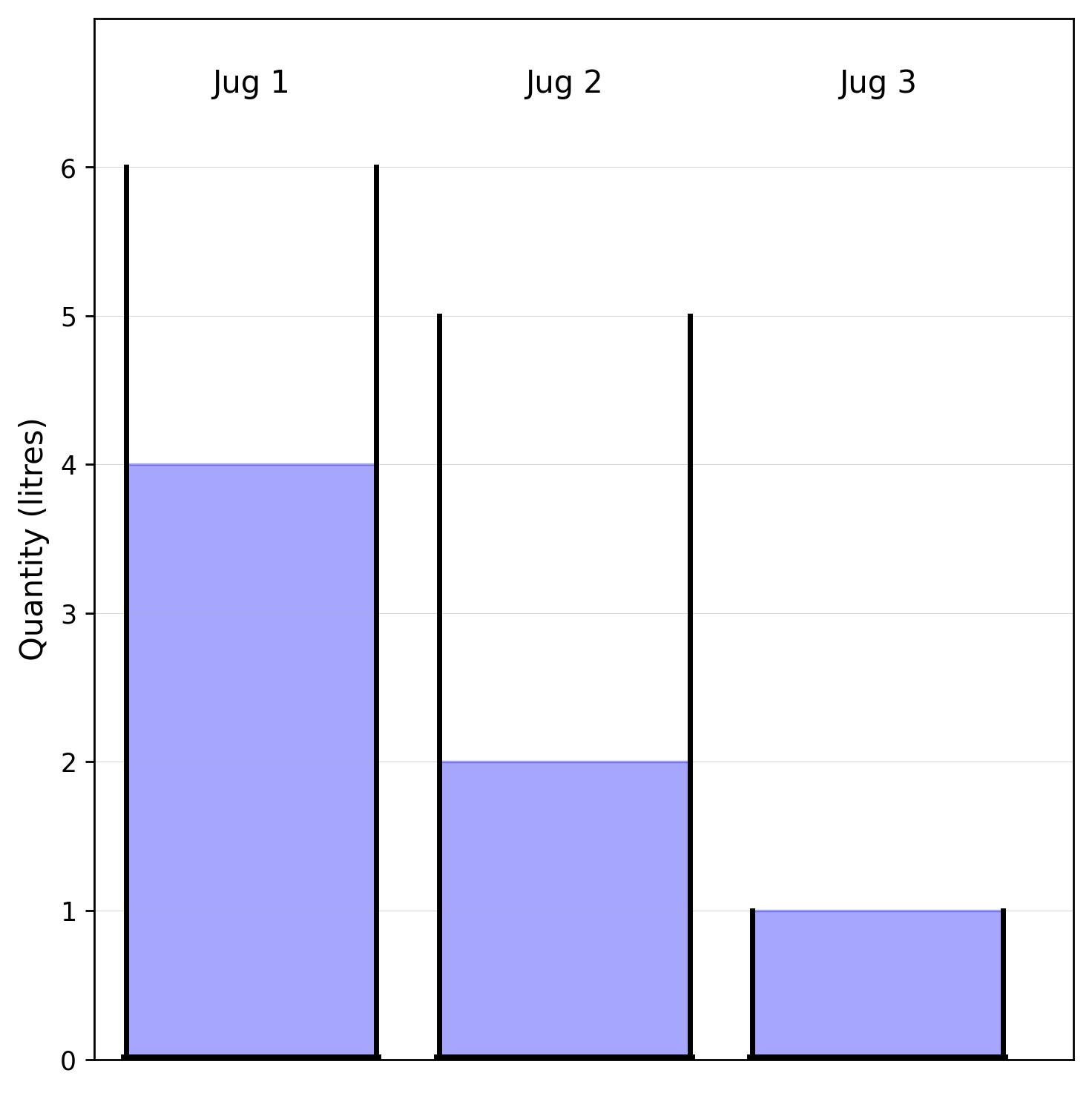} & 
The capcities, initial amount of water in each jar and objective amount of water in each jar are given in the following lists.\newline
Capacities: \texttt{[6, 5, 1]}\newline
Initial: \texttt{[4, 2, 1]}\newline
Objective: \texttt{[4, 3, 0]}
& & \\
\bottomrule
\end{tabular}
\caption{Examples of how images from different classes are converted into textual representations. In the multimodal setting, images are provided to the model, while in the text-only setting, their textual descriptions are used.}
\label{table:image-text-conversion}
\end{table*}

\begin{table*}[h!]
\centering
\small
\begin{tabular}{L{3.5cm}L{4.3cm}L{6.7cm}}
\toprule
\textbf{Image} & \textbf{Question} & \textbf{Reasoning, Answer, and Comment} \\
\midrule

\includegraphics[width=\linewidth]{figures/clock_0000.jpg} &Question: Daniel's event is going to start in 1 hour 40 minutes. The current time is shown on the clock. The clock is a standard analog clock without the
second hand. What will be the time when the event starts? \newline
\textit{Ground-Truth:} \texttt{4:23}
& 
\textit{\textbf{Reasoning}}: The clock shows an analog face with hour and minute hands. \textcolor{red}{The minute hand (brown) is pointing to 3, which indicates 15 minutes past the hour. The hour hand (green) is between 8 and 9, but closer to 8, confirming it's 8:15}. To find when the event starts, I need to add 1 hour and 40 minutes to 8:15. Adding 1 hour gives 9:15, and adding 40 more minutes results in 9:55. \newline
\textit{\textbf{Answer}}: \colorbox{red}{\textcolor{white}{9:55}} \newline 

\textit{\textbf{Comment}}: \texttt{Incorrectly interpret the positions of the minute and hour hands, leading to interpreting the wrong time. But the calculation of time is correct.} \\
\midrule

\includegraphics[width=\linewidth]{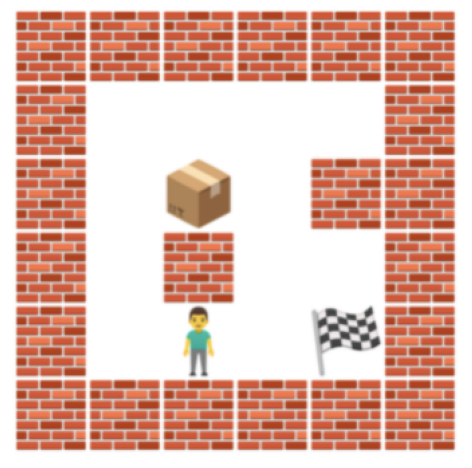} & Question: A storekeeper is a puzzle in which the player pushes boxes around in a warehouse trying to get them to target locations. The game is represented by an 6 x 6 grid of characters grid where each element is a wall, floor, or box. Your task is to move the box to the end flag under the following rules: 1. The box can be moved to an adjacent free cell by standing next to the box and then moving in the direction of the box by 1 grid. This is a push. 2. The player cannot walk through the box. What is the minimum number of pushes to move the box to the end flag. \newline
\textit{Ground-Truth:} \texttt{4} &
\textit{\textbf{Reasoning}}: The puzzle shows a maze with brick walls, a box in the center, and a checkered flag on the right. \textcolor{red}{The player starts at the bottom center}.\newline

To solve it with the fewest pushes, the player first moves up to reach the box. Then, based on the layout, the box should be pushed to the right toward the flag. \textcolor{red}{The box is about two grid spaces away, with no obstacles requiring extra moves.
Therefore, the minimum number of pushes needed is 2.} \newline
\textit{\textbf{Answer}}: \colorbox{red}{\textcolor{white}{2}} \newline 

\textit{\textbf{Comment}}: \texttt{Incorrect recognition of the spatial positions in the grid, leading to misidentification of the positions of obstacles, boxes, and the flag.} \\
\midrule
\includegraphics[width=\linewidth]{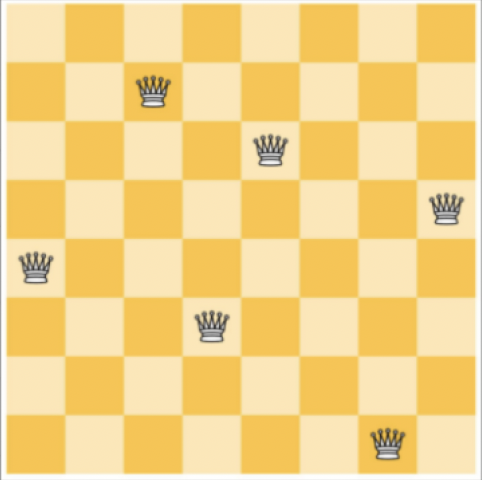} & Question:  You are given a 8 * 8 chessboard. The Manhattan distance between two squares in a chessboard is equal to the minimal number of orthogonal King moves between these squares on the otherwise empty board. The objective is to place 8 chess queens on this board so that no two queens threaten each other; i.e. no two queens share the same row, column, or diagonal. 6 queens have already been placed in some of the squares of the board, as shown in the image. Suppose you pick two squares to place the two remaining queen pieces in a way that fulfills the objective. What is the Manhattan distance between these two squares?  \newline
\textit{Ground-Truth:} \texttt{10} &
\textit{\textbf{Reasoning}}: Looking at the 8×8 chessboard, I see six queens already placed. I’ll use coordinates where (1, 1) is the top-left and (8, 8) is the bottom-right. The six queens are at:\newline
- (2, 3) – 2nd row, 3rd column \newline
\textcolor{red}{- (3, 6) – 3rd row, 6th column} \newline
- (4, 8) – 4th row, 8th column \newline
- (5, 1) – 5th row, 1st column \newline
- (6, 4) – 6th row, 4th column \newline
- (8, 7) – 8th row, 7th column \newline

To place the remaining two queens, I must find squares that are not in the same row, column, or diagonal as any existing queen. \textcolor{red}{After marking all threatened squares, only two valid positions remain: (1, 5) and (7, 2)}. The Manhattan distance between them is: |1-7|+|5-2|=6+3=9 \newline
\textit{\textbf{Answer}}: \colorbox{red}{\textcolor{white}{9}} \newline 

\textit{\textbf{Comment}}: \texttt{The queens' positions on the board were misidentified, which resulted in a wrong answer. The subsequent reasoning and calculations are accurate.} \\
\bottomrule
\end{tabular}
\caption{Qualitative example illustrating failure in visual perception leading to incorrect model predictions. Incorrect visual generations are highlighted in \textcolor{red}{red}, with our comments on the visual errors provided.}
\label{table:bottleneck-errors}
\end{table*}

\begin{figure*}[h!]
  \centering
  \includegraphics[width =\linewidth]{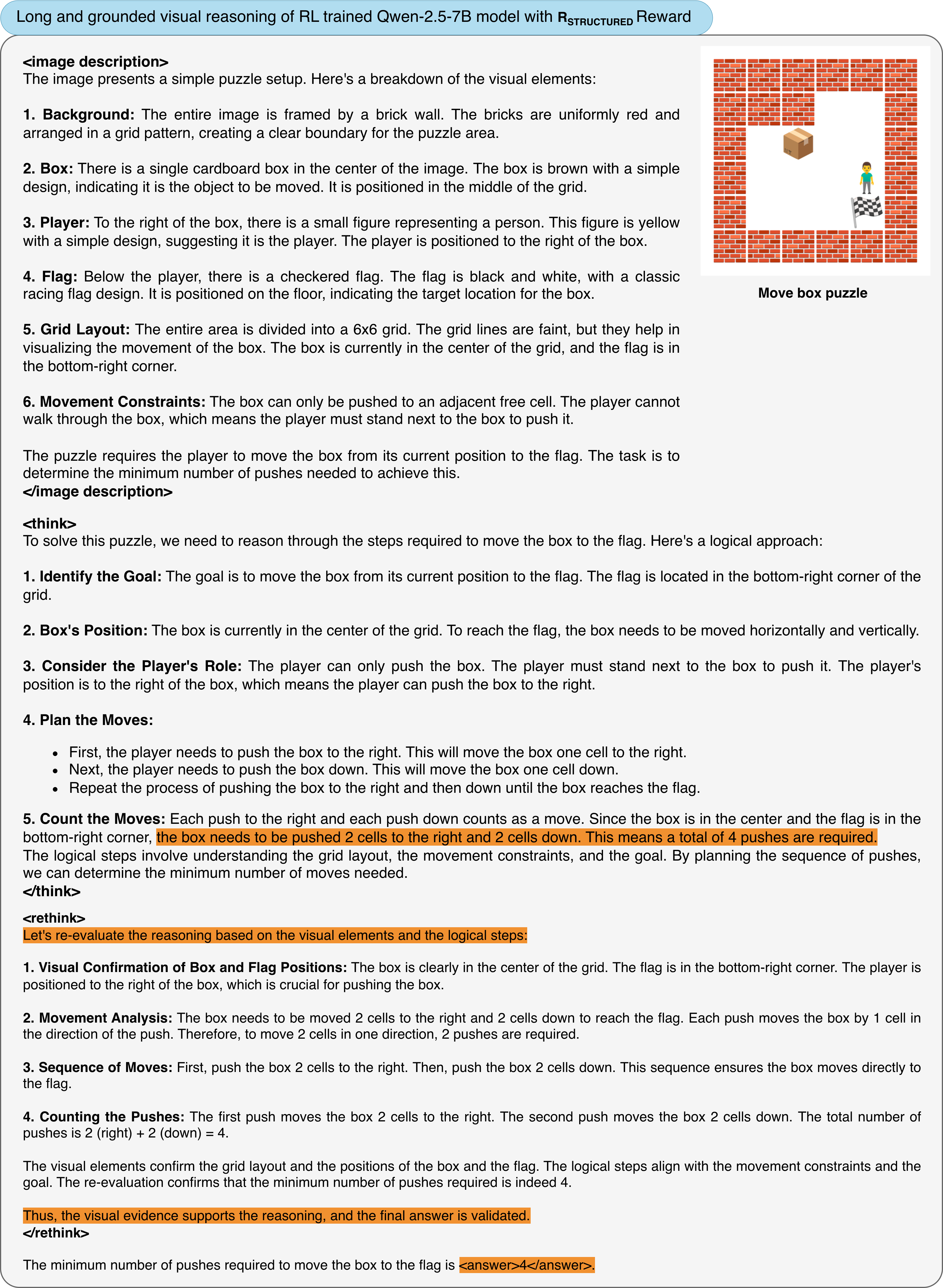}
 \caption{The RL-trained model generates detailed and well-structured reasoning for solving the puzzle, thoroughly describing all relevant visual and algorithmic elements without unnecessary repetition. This example illustrates how the \rStructured reward function guides the model toward clear, precise, and long multistep reasoning. The key part of the reasoning is bold or highlighted.}
 \label{reasoning-move-box}
\end{figure*}

\begin{figure*}[h!]
  \centering
  \includegraphics[width =\linewidth]{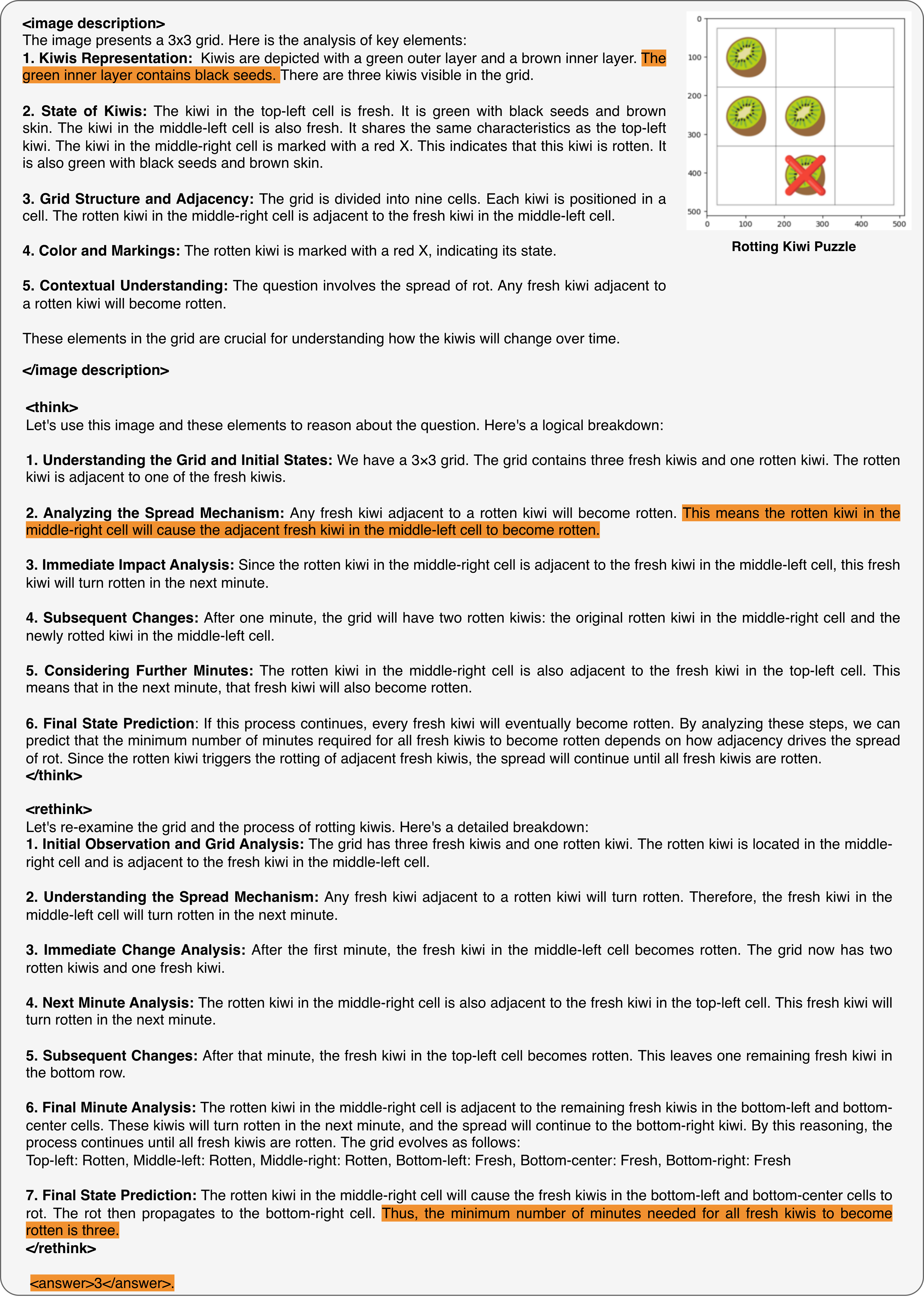}
 \caption{The RL-trained model demonstrates structured multistep visual reasoning without any supervised training samples. The reward incentivizes the model to elicit long reasoning and to explain the image in detail, like Kiwi's representation, grid structure, color, and markings. The designed reward encourages long, detailed chains of thought, yielding clear, stepwise, and self-checked explanations.}
 \label{reasoning-rotting-kiwi}
\end{figure*}

\begin{table*}[h]
\centering
\begin{tabular}{L{15cm}}

\begin{tcolorbox}[
    colback=cyan!10,
    colframe=black!90,
    title=\textbf{Prompt for \rFormat reward},
    fonttitle=\bfseries,
    sharp corners,
    boxrule=0.8pt
]
A conversation between User and Assistant. The user asks a question, and the Assistant solves it. The assistant first thinks about the reasoning process and then provides the user with the answer.\par

The reasoning process and answer are enclosed within \texttt{<think> </think>} and \texttt{<answer> </answer>} tags, respectively, i.e., <think> reasoning process here </think><answer> answer here </answer>. 
\end{tcolorbox}\\

 \begin{tcolorbox}[
    colback=cyan!10,
  colframe=black!90,
    title=\textbf{Prompt for \rOutcome, \rStructured, \rGraded, \rProcess rewards},
    fonttitle=\bfseries,
    sharp corners,
    boxrule=0.8pt
]
You are a visual reasoning model. Your task is to analyze a visual input and answer the question based on it. Follow this four-stage reasoning framework, using information from each step to inform the next:\par

\texttt{<image description>} \newline 
Begin by providing a description of the key visual elements in the image that can be relevant to the
question. Focus on objects, spatial relationships, color, text, or any other patterns that might be important.
\par
\texttt{</image description>}
\par
\texttt{<think>}
\newline
Use the image description to reason about the question. What logical steps or background knowledge help interpret the image in context of the question? Lay out general reasoning steps to solve the question without jumping to the final
answer. 
\newline
\texttt{</think>}

\texttt{<rethink>}
\newline
Reflect on the image description and the reasoning steps. Rethink and reevaluate whether the visual evidence supports the reasoning. Describe your reevaluation process. Correct or adjust your interpretation to better justify a confident final answer.
\newline
\texttt{</rethink>}
\newline
\texttt{<answer>}
\newline
Provide final answer of the question based on your validated reasoning.\newline
\texttt{</answer>}
\end{tcolorbox}
\\

\begin{tcolorbox}[
    colback=cyan!10,
   colframe=black!90,
    title=\textbf{Prompt for \rGrounded reward},
    fonttitle=\bfseries,
    sharp corners,
    boxrule=0.8pt
]
You are a visual reasoning model. Analyze an image and answer the question using structured reasoning. Begin with a
reasoning process enclosed in \texttt{<think> </think>} tags. This should reflect your internal monologue, laying out the logic step by step to answer the question. \par

Within the <think> tag, include \texttt{<visual> </visual>} tags to describe distinct and meaningful observations from the image. Each \texttt{<visual>} tag must present unique insights—do not repeat content or include vague or empty descriptions. Refine or adjust your reasoning as needed based on what you observe. At the end, provide your final answer in a single \texttt{<answer> </answer>} tag placed outside the \texttt{<think>} block. The response must include at least
one \texttt{<visual>} tag within \texttt{<think>}, and exactly one \texttt{<answer>} tag.
\end{tcolorbox}
\\

\end{tabular}
\caption{Prompts for different reward functions. We use the same prompt for the \rOutcome, \rStructured, \rGraded, and \rProcess rewards, with each function differing only in how the reward is computed. The \rFormat and \rGrounded rewards use separate prompts. All prompts were selected based on trial-and-error and our empirical observations. }
\label{table:prompts}
\end{table*}

\clearpage
\onecolumn

\begin{center}
\small

\begin{longtable}{p{0.95\textwidth}}
\begin{tcolorbox}[
    colback=gray!5,
   colframe=teal!80,
    title=\textbf{Pseudo-code for calculating \rFormat and \rOutcome rewards},
    fonttitle=\bfseries,
    sharp corners,
    boxrule=0.8pt
]
\begin{lstlisting}
FUNCTION GradeAnswer(prediction, ground_truth):
    EXTRACT all contents inside <answer> ... </answer>
    IF no answer found:
        RETURN 0
    LET final_answer be the last extracted answer
    IF final_answer matches ground_truth:
        RETURN 1
    ELSE:
        RETURN 0


FUNCTION RewardCalc(prediction, ground_truth, $\alpha_t$, $\alpha_a$):
    think_reward = 0
    
    EXTRACT all contents inside <think> ... </think>
    IF think content found:
        think_reward = $\alpha_t$
    
    accuracy_reward = $\alpha_a$ * GradeAnswer(prediction, ground_truth)
    
    RETURN {
        'overall': think_reward + accuracy_reward,
        'think': think_reward,
        'accuracy': accuracy_reward
    }

# Setting $\alpha_t = 0$ and $\alpha_a = 1$ yields in R_outcome reward
\end{lstlisting}
\end{tcolorbox}\\

\begin{tcolorbox}[
    colback=gray!5,
colframe=teal!80,
    title=\textbf{Pseudo-code for calculating \rStructured, \rGraded, and \rProcess rewards},
    fonttitle=\bfseries,
    sharp corners,
    boxrule=0.8pt
]
\begin{lstlisting}
FUNCTION SideReward(prediction, tag):
    EXTRACT text inside <tag> ... </tag>
    IF tag not present:
        RETURN 0
    
    SPLIT text into sentences
    COUNT unique_sentences
    COUNT total_sentences
    
    smoothed_score    <- tanh(unique_sentences)
    duplicate_score <- tanh(total_sentences - unique_sentences)
    RETRUN smooth_score - duplicate_score


FUNCTION RewardCalc(prediction, ground_truth, $\alpha_i$, $\alpha_t$, $\alpha_r$, $\alpha_a$):
    image_reward    <- SideReward(prediction, "image description")
    think_reward    <- SideReward(prediction, "think")
    rethink_reward  <- SideReward(prediction, "rethink")
    accuracy_reward <- GradeAnswer(prediction, ground_truth)
    
    overall_reward <- ($\alpha_i$ * image_reward + $\alpha_t$ * think_reward + $\alpha_r$ * rethinkg_reward + $\alpha_a$ * accuracy_reward)
    
    RETURN {
        'overall': overall_reward,
        'image': image_reward,
        'think': think_reward,
        'rethink': rethink_reward,
        'accuracy': accuracy_reward
    }

# the parameters $\alpha_i$, $\alpha_t$, $\alpha_r$, $\alpha_a$ are changed for different reward calculation.
\end{lstlisting}
\end{tcolorbox}\\

\begin{tcolorbox}[
    colback=gray!5,
  colframe=teal!80,
    title=\textbf{Additional details on calculating \rGraded reward},
    fonttitle=\bfseries,
    sharp corners,
    boxrule=0.8pt
]
\begin{lstlisting}
#Clock-based \rGraded reward
FUNCTION HourScore(pred_hour, true_hour):
    hour_diff = circular distance on 12-hour clock
    IF hour_diff > 2:
        RETURN 0.0
    ELSE:
        RETURN 1 - (hour_diff / 5)
    
FUNCTION MinuteScore(pred_min, true_min):
    minute_diff = circular distance on 60-minute clock
    IF minute_diff > 10:
        RETURN 0.0
    ELSE:
        RETURN 1 - (minute_diff / 20)
    
FUNCTION ClockReward(pred_time, true_time):
    CONVERT pred_time to (pred_hour, pred_minute)
    CONVERT true_time to (true_hour, true_minute)
    
    hour_score   = HourScore(pred_hour, true_hour)
    minute_score = MinuteScore(pred_minute, true_minute)
    
    RETURN 0.5 * hour_score + 0.5 * minute_score

#Numeric \rGraded reward
FUNCTION NumericReward(pred_value, true_value):
    IF conversion to number fails:
        RETURN 0.0
    numeric_diff = abs(pred_value, true_value)
    
    #if the deviation exceeds this threshold, set the reward to 0. We set it to 5.
    IF numeric_dff > 5.0:
        RETURN 0.0  
    ELSE:
        1.0/(1.0+numeric_diff) #if no error, then get full reward; otherwise, get a reward score that will linearly decay from 0.5.
    
FUNCTION GradeAnswer(prediction, ground_truth):
    EXTRACT all contents inside <answer> ... </answer>
    IF no answer found:
        RETURN 0.0
    
    LET given_answer be the last extracted answer
    
    IF given_answer and ground_truth are in time format (HH:MM):
        RETURN ClockReward(given_answer, ground_truth)
    ELSE:
        RETURN NumericReward(given_answer, ground_truth)

FUNCTION RewardCalc(prediction, ground_truth, $\alpha_i$, $\alpha_t$, $\alpha_r$, $\alpha_a$):
    image_reward    <- SideReward(prediction, "image description")
    think_reward    <- SideReward(prediction, "think")
    rethink_reward  <- SideReward(prediction, "rethink")
    continuious_reward <- GradeAnswer(prediction, ground_truth)
    
    overall_reward <- ($\alpha_i$ * image_reward + $\alpha_t$ * think_reward + $\alpha_r$ * rethinkg_reward + $\alpha_a$ * \rGraded_reward)
    
    RETURN {
        'overall': overall_reward,
        'image': image_reward,
        'think': think_reward,
        'rethink': rethink_reward,
        'accuracy': continuious_reward
    }
\end{lstlisting}
\end{tcolorbox}\\

\begin{tcolorbox}[
    colback=gray!5,
 colframe=teal!80,
    title=\textbf{Pseudo-code for calculating \rGrounded reward},
    fonttitle=\bfseries,
    sharp corners,
    boxrule=0.8pt
]
\begin{lstlisting}
FUNCTION CountUniqueVisualTags(prediction):
    EXTRACT content inside <think>
    FIND all <visual> ... </visual> tags
    REMOVE empty or duplicate visual descriptions
    COUNT unique_visual_tags
    COUNT duplicate_visual_tags
    RETURN tanh(unique_visual_tags) - tanh(duplicate_visual_tags)

FUNCTION \rGroundedReward(prediction, ground_truth, $\alpha_t$, $\alpha_v$, $\alpha_a$):
    think_reward = 0
    EXTRACT all contents inside <think> ... </think>
    IF think content found:
        think_reward = $\alpha_t$
   
    vision_reward   <- CountUniqueVisualTags(prediction)
    accuracy_reward <- GradeAnswer(prediction, ground_truth)
    
    overall_reward <- ($\alpha_v$ * vision_reward + $\alpha_t$ * think_reward + $\alpha_a$ * accuracy_reward)
    
    RETURN {
        'overall': overall_reward,
        'vision': vision_reward,
        'think': think_reward,
        'accuracy': accuracy_reward
    }

\end{lstlisting}
\end{tcolorbox}\\

\caption{Pseudo-code for reward design used in the reinforcement learning framework.
The above codes summarize the reward formulations used in our experiments. The top block shows the \rFormat and \rOutcome rewards, where answer correctness is computed by GradeAnswer and formatted reasoning is optionally encouraged via a think-token reward weighted by $\alpha_t$ (setting $\alpha_t$=0 yields \rOutcome). The middle block presents \rStructured, \rGraded, and \rProcess rewards that combine signals from image descriptions, reasoning \texttt{(<think>)}, revisions \texttt{(<rethink>)}, and answer accuracy using weights $\alpha_i, \alpha_t, \alpha_r, \alpha_a$. The bottom block details the \rGraded and \rGrounded rewards. \rGraded reward assigns partial credit based on clock-time or numeric deviation, and visual reward incentivizes grounded visual references via unique \texttt{<visual>} tags.}
\label{table:code-sample}
\end{longtable}
\end{center}

\end{document}